\documentclass[10pt]{article} % For LaTeX2e
\usepackage[accepted]{tmlr}
\usepackage{amsmath,amsfonts,bm}

\def\eqref#1{equation~\ref{#1}}
\def\1{\bm{1}}

\DeclareMathAlphabet{\mathsfit}{\encodingdefault}{\sfdefault}{m}{sl}
\SetMathAlphabet{\mathsfit}{bold}{\encodingdefault}{\sfdefault}{bx}{n}

\DeclareMathOperator*{\argmin}{arg\,min}

\usepackage[utf8]{inputenc} % allow utf-8 input
\usepackage[T1]{fontenc}    % use 8-bit T1 fonts
\usepackage{hyperref}       % hyperlinks
\usepackage{url}            % simple URL typesetting
\usepackage{booktabs}       % professional-quality tables
\usepackage{amsfonts}       % blackboard math symbols
\usepackage{nicefrac}       % compact symbols for 1/2, etc.
\usepackage{microtype}      % microtypography
\usepackage{xcolor}         % colors
\usepackage{subcaption}
\usepackage{amsmath}
\usepackage{amssymb}
\usepackage{mathtools}
\usepackage{amsthm}
\usepackage{xcolor}
\usepackage{multirow}
\usepackage{multicol}
\usepackage{graphicx}
\usepackage{algorithm,amsfonts,amsmath,amssymb,bm,booktabs,enumerate,graphicx,hyperref,microtype,nicefrac,url,wrapfig,xcolor}
\usepackage{algorithmic}

\title{Meta-Calibration: Learning of Model Calibration\\ Using Differentiable Expected Calibration Error}

\author{\name Ondrej Bohdal \email ondrej.bohdal@ed.ac.uk \\
      \addr The University of Edinburgh
      \ANDAUTHOR
      \name Yongxin Yang \email yongxin.yang@qmul.ac.uk \\
      \addr Queen Mary University of London
      \ANDAUTHOR
      \name Timothy Hospedales \email t.hospedales@ed.ac.uk \\
      \addr The University of Edinburgh \\
      Samsung AI Center Cambridge
    }

\def\month{08}  % Insert correct month for camera-ready version
\def\year{2023} % Insert correct year for camera-ready version
\def\openreview{\url{https://openreview.net/forum?id=R2hUure38l}} % Insert correct link to OpenReview for camera-ready version

\begin{document}

\maketitle

\begin{abstract}
Calibration of neural networks is a topical problem that is becoming more and more important as neural networks increasingly underpin real-world applications. The problem is especially noticeable when using modern neural networks, for which there is a significant difference between the confidence of the model and the probability of correct prediction. Various strategies have been proposed to improve calibration, yet accurate calibration remains challenging. We propose a novel framework with two contributions: introducing a new differentiable surrogate for expected calibration error (DECE) that allows calibration quality to be directly optimised, and a meta-learning framework that uses DECE to optimise for validation set calibration with respect to model hyper-parameters. The results show that we achieve competitive performance with existing calibration approaches. Our framework opens up a new avenue and toolset for tackling calibration, which we believe will inspire further work on this important challenge.
\end{abstract}

\section{Introduction}
When deploying neural networks to real-world applications, it is crucial that models' 
own confidence estimates accurately match their probability of making a correct prediction. If a model is over-confident about its predictions, we cannot rely on it; while well-calibrated models can abstain or ask for human feedback in the case of uncertain predictions. Models with accurate confidence estimates about their own predictions can be described as well-calibrated. This is particularly important in applications involving safety or human impact -- such as autonomous vehicles \citep{Bojarski2016EndCars,wiseman2022cars} and medical diagnosis \citep{Jiang2012CalibratingMedicine, Caruana2015IntelligibleHealthcare, ElSappagh2023TrustworthyAI}, and tasks that directly rely on calibration such as outlier detection \citep{Hendrycks2017ANetworks,Liang2018EnhancingNetworks, Wang2023OutofdistributionDW}. However, modern neural networks are known to be badly calibrated \citep{Guo2017OnNetworks,Gawlikowski2021ASO}. 

This challenge of calibrating neural networks has motivated a growing area of research. Perhaps the simplest approach is to post-process predictions with techniques such as temperature scaling \citep{Guo2017OnNetworks}. However, this has limited efficacy \citep{wang2021rethinking} and fails in the common situation of distribution shift between training and testing data \citep{ovadia2019trustUncertainty,tomani2021calibrationDomain}. It also reduces network's confidence in correct predictions. Another family of approaches modifies the model training regime to improve calibration. \citet{Muller2019WhenHelp} show that label smoothing regularization improves calibration by increasing overall predictive entropy. But it is unclear how to set the label smoothing parameter so as to optimise calibration. 
\citet{Mukhoti2020CalibratingLoss} show that Focal loss leads to better calibrated models than standard cross-entropy, and \citet{Kumar2018TrainableEmbeddings} introduce a proxy for calibration error to be minimised along with standard cross-entropy on the training set. However, this does not ensure calibration on the test set. \citet{ovadia2019trustUncertainty,mosser2022calibration} show that Bayesian neural network approaches are comparatively well-calibrated, however these are difficult and expensive to scale to large modern architectures.

\begin{figure}[t]
\vskip 0.1in
\begin{center}
\centerline{\includegraphics[width=0.8\columnwidth]{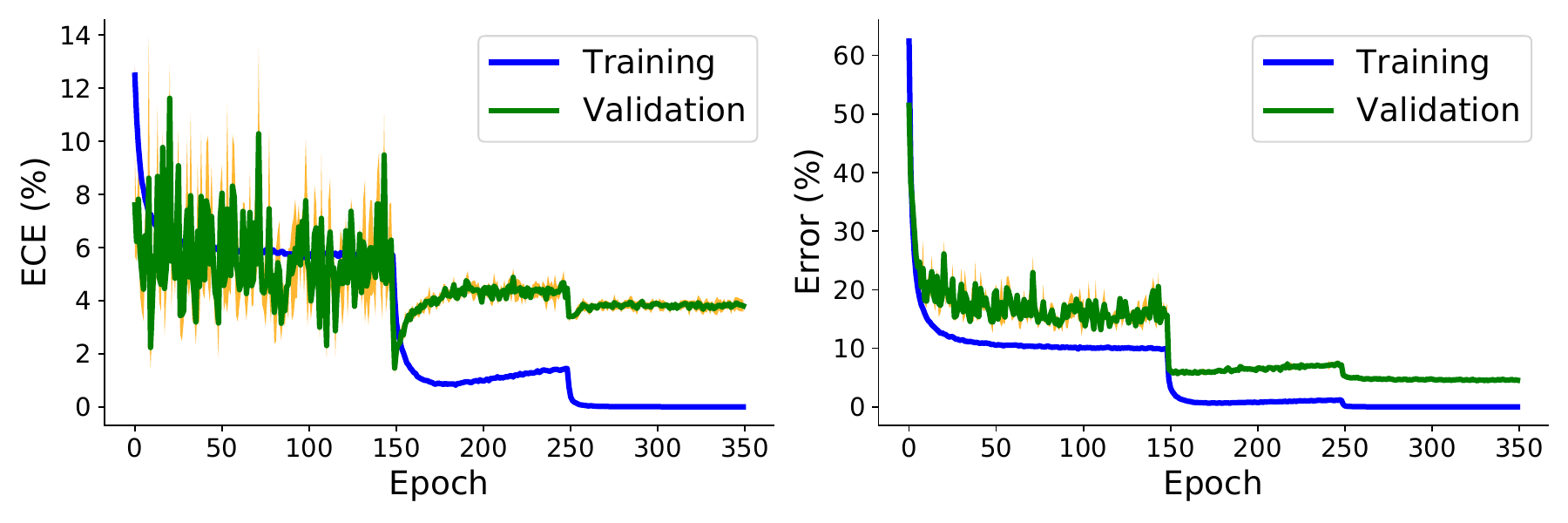}}
\caption{Illustration of the calibration generalisation gap. ECE and classification error during training of ResNet18 on CIFAR-10, using cross-entropy loss and showing mean and std. across three random seeds. Training ECE and error fall to 0. However, calibration overfitting occurs and validation ECE increases. This motivates the need for meta-learning to tune hyper-parameters to optimise \emph{validation} calibration.}
\label{fig:ece-error-behaviour}
\end{center}
\vskip -0.1in
\end{figure}

The above approaches explore the impact of various architectures and design parameters on calibration. In this paper we step back and consider how to optimise for calibration. Direct optimisation for calibration would require a differentiable metric for calibration. However, calibration is typically measured using expected calibration error (ECE), which is not differentiable due to its internal binning/counting operation. Therefore our first contribution is a high-quality differentiable approximation to ECE, which we denote DECE. A second consideration is how to optimise for calibration -- given that calibration itself is a quantity with a generalisation gap between training and validation \citep{carrell2022calibration}. We illustrate this challenge in  Figure~\ref{fig:ece-error-behaviour}, which shows how validation calibration worsens as training calibration improves during training. 
To this end, our second contribution is to introduce a framework for meta-learning model calibration: We fit a model on the training set using a given set of hyper-parameters, evaluate it on a disjoint validation set, and optimise for the hyper-parameters that lead to the best \emph{validation} calibration as measured by DECE.
Our framework for differentiable optimisation of validation calibration is generic and can potentially be used with any continuous model hyper-parameters. Our third contribution is a specific choice of hyper-parameters which, when meta-learned with a suitable calibration objective, are effective for tuning the base model's calibration. Specifically, we propose non-uniform label-smoothing, which can be tuned by meta-learning to penalise differently each unique combination of true and predicted label.

To summarize, we present a novel framework and toolset for improving model calibration by differentiable optimisation of model hyper-parameters with respect to validation calibration. We analyse our differentiable calibration metric in detail, and show that it closely matches the original non-differentiable metric. When instantiated with label smoothing hyper-parameters, our empirical results show that our framework produces high-accuracy and well-calibrated models that are competitive with existing methods across a range of benchmarks and architectures. 

\section{Related work}

\paragraph{Calibration}

Since finding modern neural networks are typically miscalibrated \citep{Guo2017OnNetworks}, model calibration has become a popular area for research \citep{Gawlikowski2021ASO} with many applications, including medical image segmentation \citep{Judge2022CRISPR} and object detection \citep{Munir2023BridgingPA}. \citet{Guo2017OnNetworks} study a variety of potential solutions and find simple post-training rescaling of the logits -- temperature scaling -- works relatively well. \citet{Kumar2018TrainableEmbeddings} propose a kernel-based measure of calibration called MMCE that they use as regularization during training of neural networks. \citet{Mukhoti2020CalibratingLoss} show Focal loss -- a relatively simple weighted alternative to cross-entropy -- can be used to train well-calibrated neural networks. The classic Brier score \citep{Brier1950VerificationProbability}, which is the squared error between the softmax vector with probabilities and the ground-truth one-hot encoding, has also been shown to work well. Similarly, label smoothing \citep{Muller2019WhenHelp} has been shown to improve model calibration. Noise during training more broadly can also be beneficial for improving calibration \citep{Ferianc2023ImpactNetworks}. These aforementioned methods do not optimise for calibration metric (e.g., ECE) directly, because the calibration metrics are usually non-differentiable. In this work, we propose a new high-quality differentiable approximation to ECE, and utilize it with meta-learning.

\citet{Karandikar2021SoftNetworks} have proposed soft-binned ECE (SB-ECE) as an auxiliary loss to be used during training to encourage better calibration. The approach makes the binning operation used in ECE differentiable, leading to an additional objective that is more compatible with gradient-based methods. SB-ECE does not make the accuracy component of ECE differentiable, but we make all components of ECE differentiable. We also try to obtain a highly accurate approximation of the binning operation, while SB-ECE binning estimate for the left-most and right-most bin can be inaccurate as a result of using bin's center value. We provide additional comparison with SB-ECE in the appendix, showing DECE provides a closer approximation to ECE than SB-ECE, and that empirically our meta-learning approach with DECE leads to better calibration.

\paragraph{Meta-learning}
We use the newly introduced DECE metric as part of the meta-objective for gradient-based meta-learning. Gradient-based meta-learning has become popular since the seminal work MAML \citep{Finn2017Model-agnosticNetworks} has successfully used it to solve the challenging problem of few-shot learning \citep{Wang2020GeneralizingLearning,Song2023AOpportunities,Bohdal2023MetaLearning-to-learn}. Nevertheless, gradient-based meta-learning is not limited to few-shot learning problems, but it can also be used to solve various other challenges, including training with noisy labels \citep{Shu2019Meta-Weight-Net:Weighting,algan2022noisy}, dataset distillation \citep{Wang2018DatasetDistillation, Bohdal2020FlexibleImages,Yu2023DatasetDA}, domain generalization and adaptation \citep{Li2019Feature-criticGeneralization, Balaji2018MetaReg:Meta-regularization,Bohdal2022Feed-forwardCross-attention}, molecular property prediction \citep{Chen2022MetalearningAD} and many others.

Gradient-based meta-learning is typically formulated as a bilevel optimisation problem where the main model is trained in the inner loop and the meta-knowledge or hyper-parameters are trained in the outer loop. In the case of few-shot learning it is possible to fully train the model within the inner loop -- also known as offline meta-learning. In more realistic and larger scale settings such as ours, it is only feasible to do one or a few updates in the inner loop. This approach is known as online meta-learning \citep{Hospedales2021Meta-learningSurvey} and means that we jointly train the main model as well as the meta-knowledge. Online meta-learning is most commonly done using the so-called $T_1-T_2$ method \citep{Luketina2016ScalableHyperparameters} that updates the meta-knowledge by backpropagating through one step of the main model update. This is the approach that we adopt, however more advanced or efficient approaches are also available \citep{Lorraine2020optimizingDifferentiation, Bohdal2021EvoGrad:optimization}.

\paragraph{Label smoothing}

We use label smoothing as the meta-knowledge that we use to demonstrate the benefits of using our DECE metric. Label smoothing has been proposed by \citet{Szegedy2016RethinkingVision} as a technique to alleviate overfitting and improve the generalization of neural networks. It consists of replacing the one-hot labels by their softer alternative that gives non-zero target probabilities to the incorrect classes. Label smoothing has been studied in more detail, for example \citet{Muller2019WhenHelp} have observed that label smoothing can improve calibration, but at the same time it can hurt knowledge distillation if used for training the teacher. \citet{Mukhoti2020CalibratingLoss} have compared Focal loss with label smoothing among other approaches, showing that simple label smoothing strategy has a limited scope for state-of-the-art calibration. 

However, we demonstrate that using meta-learning and our DECE objective, a more expressive form of label smoothing can achieve state-of-the-art calibration results. Note that meta-learning has already been used for label smoothing \citep{Li2020LearningLearning}, but using it as meta-knowledge to directly optimise calibration is new.

\section{Methods}
\subsection{Preliminaries}
We first discuss expected calibration error (ECE) \citep{Naeini2015ObtainingBinning}, before we derive a differentiable approximation to it. ECE measures the expected difference (in absolute value) between the accuracies and the confidences of the model on examples that belong to different confidence intervals. ECE is defined as
$$\mathrm{ECE}=\sum_{m=1}^{M} \frac{\left|B_{m}\right|}{n}\left|\operatorname{acc}\left(B_{m}\right)-\operatorname{conf}\left(B_{m}\right)\right|,$$
where accuracy and confidence for bin $B_m$ are
\begin{align}
\operatorname{acc}\left(B_{m}\right)&=\frac{1}{\left|B_{m}\right|} \sum_{i \in B_{m}} \mathbf{1}\left(\hat{y}_{i}=y_{i}\right)\nonumber\\
\operatorname{conf}\left(B_{m}\right)&=\frac{1}{\left|B_{m}\right|} \sum_{i \in B_{m}} \hat{p}_{i}.\nonumber
\end{align}

There are $M$ interval bins each of size $1/M$ and $n$ samples. Confidence $\hat{p}_i$ is the probability of the top prediction as given by the model for example $i$. We group the confidences into their corresponding bins, with bin $B_m$ covering interval $(\frac{m-1}{M}, \frac{m}{M}]$. The predicted class of example $i$ is $\hat{y}_i$, while $y_i$ is the true class of example $i$ and $\mathbf{1}$ is an indicator function.

ECE metric is not differentiable because assigning examples into bins is not differentiable and also accuracy is not differentiable due to the indicator function. We propose approximations to both binning and accuracy and derive a new metric called differentiable ECE (DECE).

\subsection{Differentiable ECE}
ECE is composed of accuracy, confidence and binning, but only confidence is differentiable.

\paragraph{Differentiable accuracy} In order to obtain a differentiable approximation to accuracy, we consider approaches that allow us to find a differentiable way to calculate the rank of a given class. Two approaches stand out: differentiable ranking \citep{Blondel2020FastRanking} and an all-pairs approach \citep{Qin2010AMeasures}. While both allow us to approximate the rank in a differentiable way, differentiable ranking is implemented on CPU only, which would introduce a potential bottleneck for modern applications. All-pairs approach has asymptotic complexity of $\mathcal{O}(n^2)$ for $n$ classes, while differentiable ranking is $\mathcal{O}(n \log n)$. However, if the number of classes is not in thousands or millions, differentiable ranking would be slower because of not using GPUs. We use the all-pairs approach to estimate the rank of a given class.

All-pairs \citep{Qin2010AMeasures} calculates a rank of class $i$ as $\left[R\left(\cdot\right)\right]_i = 1+\sum_{j \neq i} \mathbf{1}\left[\boldsymbol{\phi}_{i}<\boldsymbol{\phi}_{j}\right],$ where $\boldsymbol{\phi}$ are the logits. We can obtain soft ranks by replacing the indicator function with a sigmoid scaled with some temperature value $\tau_a$ to obtain reliable estimates of the rank of the top predicted class. Once the rank $[R(\cdot)]_l$ for true class $l$ is calculated, we can estimate the accuracy as $\textrm{acc} = \max(0, 2-[R]_l)$.

\paragraph{Soft binning} Our approach is similar to \citep{Yang2018DeepTrees}. We take confidence $\hat{p}_i$ for example $x_i$ and pass it through one-layer neural network $\text{softmax}((w\hat{p}_i+b)/\tau_b)$ parameterized with different values of $w$ and $b$ as explained in \citep{Yang2018DeepTrees}, with temperature $\tau_b$ to control the binning. This leads to $M$ different probabilities, saying how likely it is that $\hat{p}_i$ belongs to the specific bin $B_{m \in 1..M}$. We will denote these probabilities as $o_m(x_i) = p (B_m|\hat{p}_i)$.

Putting these parts together, we define DECE using a minibatch of $n$ examples as:
\begin{align}
   \mathrm{DECE}&=\sum_{m=1}^{M} \frac{\sum_{i=1}^n o_m(x_i)}{n}\left|\operatorname{acc}\left(B_{m}\right)-\operatorname{conf}\left(B_{m}\right)\right|,\nonumber\\
\operatorname{acc}\left(B_{m}\right)&=\frac{1}{\sum_{i=1}^n o_m(x_i)} \sum_{i=1}^n o_m(x_i) \mathbf{1}\left(\hat{y}_{i}=y_{i}\right),\nonumber\\
\operatorname{conf}\left(B_{m}\right)&=\frac{1}{\sum_{i=1}^n o_m(x_i)} \sum_{i=1}^n o_m(x_i) \hat{p}_{i}.\nonumber
\end{align}

\subsection{Meta-learning}
Differentiable ECE provides an objective to optimise, but we still need to decide how to utilize it. One option could be to directly use it as an extra objective in combination with standard cross-entropy, as used by a few existing attempts \citep{Karandikar2021SoftNetworks,Kumar2018TrainableEmbeddings}. However, we expect this to be unhelpful as calibration on the \emph{training} set is usually good -- the issue being a failure of calibration generalisation to the held out validation or test set \citep{carrell2022calibration}, as illustrated in Figure~\ref{fig:ece-error-behaviour}. Meanwhile multi-task training with such non-standard losses may negatively affect the learning dynamics of existing well tuned model training procedures. To optimise for calibration of held-out data, without disturbing standard model training dynamics, we explore the novel approach of using DECE as part of the objective for hyper-parameter meta-learning in an outer loop that wraps an inner learning process of conventional cross-entropy-driven model training.

\paragraph{Meta-learning objective}
We formulate our approach as a bilevel optimisation problem. Our model is assumed to be composed of feature extractor $\boldsymbol{\theta}$ and classifier $\boldsymbol{\phi}$. These are trained to minimise $\mathcal{L}^{train}_{CE}$, cross-entropy loss on the training set. The goal of meta-learning is to find hyper-parameters $\boldsymbol{\omega}$ so that training with them optimises the meta-objective computed on the meta-validation set. In our case the meta-objective is a combination of cross-entropy and DECE to reflect that the meta-learned hyper-parameters should lead to both good generalization and calibration. More specifically the meta-objective is $\mathcal{L}^{val}_{CE + \lambda DECE}$, with hyper-parameter $\lambda$ specifying how much weight is placed on calibration. The bilevel optimisation problem can then be summarized as:
\begin{align}\label{eq:BLO}
\boldsymbol{\omega}^* &= \argmin_{\boldsymbol{\omega}}\mathcal{L}^{val}_{CE + \lambda DECE}(\boldsymbol{\phi}^*\circ\boldsymbol{\theta}^*(\boldsymbol{\omega})),\nonumber\\ \boldsymbol{\phi}^*,\boldsymbol{\theta}^*(\boldsymbol{\omega})&=\argmin_{\boldsymbol{\phi}, \boldsymbol{\theta}} \mathcal{L}^{train}_{CE}\left(\boldsymbol{\phi} \circ \boldsymbol{\theta}, \boldsymbol{\omega}\right).
\end{align}

To solve this efficiently, we adopt online meta-learning approach \citep{Luketina2016ScalableHyperparameters,Hospedales2021Meta-learningSurvey} where we alternate base model and hyper-parameter updates. This is an efficient strategy as we do not need to backpropagate through many inner-loop steps or retrain the model from scratch for each update of meta-knowledge.

When simulating training during the inner loop, we only update the classifier and keep the feature extractor frozen for efficiency, as suggested by \citep{Balaji2018MetaReg:Meta-regularization}.
Base model training is done separately using a full model update and a more advanced optimiser. 

We give the overview of our meta-learning algorithm in Algorithm \ref{alg:onlinemetalearning}. The inner loop that trains the main model $(\boldsymbol{\theta},\boldsymbol{\phi})$ (line 10) is conducted using hyper-parameters $\boldsymbol{\omega}$, while the outer loop (line 12) that trains the hyper-parameters does not directly use it for evaluating the meta-objective (e.g. no learnable label smoothing is applied to the meta-validation labels that are used in the outer loop). We backpropagate through one step of update of the main model.

\begin{algorithm}[t]
   \caption{Meta-Calibration}
   \label{alg:onlinemetalearning}
\begin{algorithmic}[1]
   \STATE {\bfseries Input:} $\alpha$, $\beta$: inner and outer-loop learning rates
   \STATE {\bfseries Output:} trained feature extractor $\boldsymbol{\theta}$, classifier $\boldsymbol{\phi}$ and label smoothing $\boldsymbol{\omega}$
   \STATE $\boldsymbol{\omega} \sim p(\boldsymbol{\omega})$
   \STATE $\boldsymbol{\phi}, \boldsymbol{\theta} \sim p(\boldsymbol{\phi}), p(\boldsymbol{\theta})$
   \WHILE{training}
   \STATE Sample minibatch of training $x_t, y_t$ and meta-validation $x_v, y_v$ examples
   \STATE \textbf{// For LS: use $\boldsymbol{\omega}$ to smooth $y_t$}
   \STATE \textbf{// For L2: add unit-wise weight-decay $\boldsymbol{\omega}$}
   \STATE Calculate $\mathcal{L}_i=\mathcal{L}_{CE}\left(f_{\boldsymbol{\phi} \circ \boldsymbol{\theta}}\left(x_t\right), y_{t}, \boldsymbol{\omega}\right)$
   \STATE Update $\boldsymbol{\theta}, \boldsymbol{\phi} \leftarrow \boldsymbol{\theta}, \boldsymbol{\phi} - \alpha\nabla_{\boldsymbol{\theta}, \boldsymbol{\phi}} \mathcal{L}_i$
   \STATE Calculate $\mathcal{L}_o=\mathcal{L}_{CE + \lambda DECE}\left(f_{\boldsymbol{\phi} \circ \boldsymbol{\theta}}\left(x_v\right), y_v\right)$
   \STATE Update $\boldsymbol{\omega} \leftarrow \boldsymbol{\omega} - \beta\nabla_{\boldsymbol{\omega}} \mathcal{L}_o$
   \ENDWHILE
\end{algorithmic}
\end{algorithm}

\paragraph{Hyper-parameter choice} 
A key part of meta-learning is to select suitable meta-knowledge (hyper-parameters) that we will optimise to
achieve the meta-learning goal \citep{Hospedales2021Meta-learningSurvey}. Having cast calibration optimisation as a meta-learning process, we are free to use any of the wide range of hyper-parameters surveyed in  \citep{Hospedales2021Meta-learningSurvey}. Note that in contrast to grid search that standard temperature scaling \citep{Guo2017OnNetworks, Mukhoti2020CalibratingLoss} and other approaches rely on, we have gradients with respect to hyper-parameters and so can therefore potentially optimise calibration with respect to high-dimensional hyper-parameters. In this paper we explore two options to demonstrate this generality: 1) Unit-wise L2 regularization coefficients of each weight in the classifier layer, inspired by \citep{Balaji2018MetaReg:Meta-regularization} and \citep{Lorraine2020optimizingDifferentiation}; and 2) various types of learnable label smoothing (LS) \citep{Muller2019WhenHelp}. However, we found LS to be better overall, so our experiments focus on this and selectively compare against L2.

\paragraph{Learnable label-smoothing} 
Learnable label smoothing learns one or more coefficients to smooth the one-hot encoded labels. More formally, if there are $K$ classes in total, $y_k$ is 1 for the correct class $k=c$ and $y_k$ is 0 for all classes $k\neq c$, then with learnable label smoothing $\boldsymbol{\omega}$ the soft label for class $k$ becomes $$y_k^{LS}=y_k(1-\omega_{c,k})+\omega_{c,k} / K.$$
In the scalar case of label smoothing, $\omega_c$ is the same for all classes, while for the vector case it takes different values for each class $c$. We consider scalar and vector variations as part of ablation.

Our main variation of meta-calibration uses non-uniform label smoothing. It is computed using the overall strength of smoothing $\omega^{s}_c$ for the correct class $c$ and also $\omega^{d}_{c,k}$ saying how $\omega^{s}_c$ is distributed across the various incorrect classes $k \neq c$. Given correct class $c$, with this variation the soft label for class $k$ is calculated as:
$$y_k^{LS}=y_k(1-\omega^{s}_c)+\omega^{s}_c\frac{\omega^{d}_{c,k}}{\epsilon + \sum_{i=1}^K \omega^{d}_{c,i}},$$

where we normalize the distribution weights to sum to 1 and use small value $\epsilon$ to avoid division by 0. The learnable label smoothing parameters are restricted to non-negative values, with the total label smoothing strength at most 0.5. In practice the above is implemented by learning a vector of $K$ elements specifying the strengths of overall label smoothing for different correct classes $c$, and a matrix of $K \times (K-1)$ elements specifying how the label smoothing is distributed across the incorrect classes $k\neq c$.

\paragraph{Learnable L2 regularization} 
In the case of learnable L2 regularization (cf: \citet{Balaji2018MetaReg:Meta-regularization}), the goal is to find unit-wise L2 regularization coefficients $\boldsymbol{\omega}$ for the classifier layer $\boldsymbol{\phi}$ so that training with them optimises the meta-objective that includes DECE ($\boldsymbol{\theta}$ is the feature extractor). The inner loop loss becomes $$\mathcal{L}_i=\mathcal{L}_{CE}\left(f_{\boldsymbol{\phi} \circ \boldsymbol{\theta}}\left(x_t\right), y_t\right) + \boldsymbol{\omega} \lVert \boldsymbol{\phi} \rVert ^2.$$

\section{Experiments}
Our experiments show that DECE-driven meta-learning can be used to obtain excellent calibration across a variety of benchmarks and models. 

\subsection{Calibration experiments}
\paragraph{Datasets and settings}
We experiment with CIFAR-10 and CIFAR-100 benchmarks \citep{Krizhevsky2009LearningImages}, SVHN \citep{Netzer2011ReadingLearning} and
20 Newsgroups dataset \citep{Lang1995Newsweeder:Netnews}, covering both computer vision and NLP. 
For CIFAR benchmark, we use ResNet18, ResNet50, ResNet110 \citep{He2015DeepRecognition} and WideResNet26-10 \citep{Zagoruyko2016WideNetworks} models. For SVHN we use ResNet18, while for 20 Newsgroups we use global pooling CNN \citep{Lin2014NetworkNetwork}. We extend the implementation provided by \citep{Mukhoti2020CalibratingLoss} to implement and evaluate our meta-learning approach. We use the same hyper-parameters as selected by the authors for fair comparison, which we summarize next.

CIFAR and SVHN models are trained for 350 epochs, with a multi-step scheduler that decreases the initial learning rate of 0.1 by a factor of 10 after 150 and 250 epochs. Each model is trained with SGD with momentum of 0.9, weight decay of 0.0005 and minibatch size of 128. 90\% of the original training set is used for training and 10\% for validation. In the case of meta-learning, we create a further separate \emph{meta-validation} set that is of size 10\% of the original training data, so we directly train with 80\% of the original training data. 20 Newsgroups models are trained with Adam optimiser with the default parameters, 128 minibatch size and for 50 epochs. As the final model we select the checkpoint with the best validation accuracy.

For DECE, we use $M=15$ bins and scaling parameters $\tau_a=100, \tau_b=0.01$. Learnable label smoothing coefficients are optimised using Adam \citep{Kingma2015Adam:optimization} optimiser with learning rate of 0.001. The meta-learnable parameters are initialized at 0.0 (no label smoothing or L2 regularization initially). The total number of meta-parameters is $K\times K$, 1 and $K$ for the non-uniform, scalar and vector label smoothing respectively, while it is $512 \times  K + K$ for learnable L2 regularization. We use $\lambda=0.5$ in the meta-objective, and we have selected it based on validation set calibration and accuracy after trying several values.

\paragraph{Results}
We first follow the experimental setup of \citet{Mukhoti2020CalibratingLoss} and compare with the following alternatives: 1) cross-entropy, 2) Brier score \citep{Brier1950VerificationProbability}, 3) Weighted MMCE \citep{Kumar2018TrainableEmbeddings} with $\lambda=2$, 4) Focal loss \citep{Lin2017FocalDetection} with $\gamma=3$, 5) Adaptive (sample dependent) focal loss  (FLSD) \citep{Mukhoti2020CalibratingLoss} with $\gamma=5$ and $\gamma=3$ for predicted probability $p\in[0, 0.2)$ and $p\in[0.2, 1)$ respectively. 6) Label smoothing (LS) with a fixed smoothing factor of 0.05. In all cases we report the mean and standard deviation across three repetitions to obtain a more reliable estimate of the performance. In contrast, \citet{Mukhoti2020CalibratingLoss} report their results on only one run, so in the tables we include our own results for the comparison with the baselines.

We show the test ECE, test ECE after temperature scaling (TS) and test error rates in Tables~\ref{tab:main_ece}, \ref{tab:main_ece_temp} and \ref{tab:main_error} respectively. 
Meta-Calibration leads to excellent intrinsic calibration without the need for post-processing (Table~\ref{tab:main_ece}), which is practically valuable because post-processing is not always possible \citep{Kim2020TaskClassification} or reliable \citep{ovadia2019trustUncertainty}. However, even after post-processing using TS Meta-Calibration gives competitive performance (Table~\ref{tab:main_ece_temp}), as evidenced by the best average rank across the considered scenarios. Table~\ref{tab:main_error} shows that Meta-Calibration maintains comparable accuracy to the competitors, even if it does not have the best average rank there. Overall Meta-Calibration leads to significantly better intrinsic calibration, while keeping similar or only marginally worse accuracy. 

\begin{table*}[h!]
\caption{Test ECE (\%, $\downarrow$): Our Meta-Calibration (MC) leads to excellent intrinsic calibration.}
\label{tab:main_ece}
\vskip 0.1in
\centering
\resizebox{1.\linewidth}{!}{\begin{tabular}{llccccccccc}
\toprule
Dataset & Model & CE & Brier & MMCE & FL-3 & FLSD-53 & LS & MC (Ours) \\
\midrule
\multirow{4}{*}{CIFAR-10} & ResNet18 & \phantom{0}4.23 $\pm$ 0.15 & \phantom{0}1.23 $\pm$ 0.03 & \phantom{0}4.36 $\pm$ 0.16 & \phantom{0}2.11 $\pm$ 0.09 & \phantom{0}2.22 $\pm$ 0.04 & \phantom{0}3.63 $\pm$ 0.06 & 1.17 $\pm$ 0.26 \\
& ResNet50 & \phantom{0}4.20 $\pm$ 0.01 & \phantom{0}1.95 $\pm$ 0.15 & \phantom{0}4.49 $\pm$ 0.18 & \phantom{0}1.48 $\pm$ 0.19 & \phantom{0}1.68 $\pm$ 0.14 & \phantom{0}2.58 $\pm$ 0.26 & 1.09 $\pm$ 0.09 \\
& ResNet110 & \phantom{0}4.81 $\pm$ 0.12 & \phantom{0}2.58 $\pm$ 0.17 & \phantom{0}4.20 $\pm$ 0.74 & \phantom{0}1.82 $\pm$ 0.20 & \phantom{0}2.16 $\pm$ 0.22 & \phantom{0}1.96 $\pm$ 0.36 & 1.07 $\pm$ 0.12 \\
& WideResNet26-10 & \phantom{0}3.37 $\pm$ 0.11 & \phantom{0}1.03 $\pm$ 0.08 & \phantom{0}3.48 $\pm$ 0.06 & \phantom{0}1.57 $\pm$ 0.32 & \phantom{0}1.50 $\pm$ 0.15 & \phantom{0}3.68 $\pm$ 0.10 & 0.94 $\pm$ 0.10 \\
\midrule
\multirow{4}{*}{CIFAR-100} & ResNet18 & \phantom{0}8.79 $\pm$ 0.59 & \phantom{0}5.19 $\pm$ 0.18 & \phantom{0}7.41 $\pm$ 1.30 & \phantom{0}2.83 $\pm$ 0.27 & \phantom{0}2.47 $\pm$ 0.12 & \phantom{0}6.87 $\pm$ 0.29 & 2.52 $\pm$ 0.35 \\
& ResNet50 & 12.56 $\pm$ 1.44 & \phantom{0}4.82 $\pm$ 0.36 & \phantom{0}9.02 $\pm$ 1.72 & \phantom{0}4.78 $\pm$ 1.00 & \phantom{0}5.43 $\pm$ 0.31 & \phantom{0}5.94 $\pm$ 0.52 & 3.07 $\pm$ 0.18 \\
& ResNet110 & 14.96 $\pm$ 0.83 & \phantom{0}6.52 $\pm$ 0.56 & 12.29 $\pm$ 1.25 & \phantom{0}6.64 $\pm$ 1.42 & \phantom{0}7.38 $\pm$ 0.25 & 10.69 $\pm$ 0.39 & 2.80 $\pm$ 0.58 \\
& WideResNet26-10 & 12.39 $\pm$ 1.44 & \phantom{0}4.26 $\pm$ 0.30 & \phantom{0}8.35 $\pm$ 2.79 & \phantom{0}2.36 $\pm$ 0.13 & \phantom{0}2.30 $\pm$ 0.36 & \phantom{0}3.94 $\pm$ 0.96 & 3.86 $\pm$ 0.34 \\
\midrule
SVHN & ResNet18 & \phantom{0}2.98 $\pm$ 0.08 & \phantom{0}1.94 $\pm$ 0.10 & \phantom{0}3.14 $\pm$ 0.10 & \phantom{0}2.69 $\pm$ 0.06 & \phantom{0}2.83 $\pm$ 0.17 & \phantom{0}3.88 $\pm$ 0.01 & 1.14 $\pm$ 0.12 \\
\midrule
20 Newsgroups & Global Pooling CNN & 18.58 $\pm$ 0.80 & 16.49 $\pm$ 0.70 & 14.68 $\pm$ 1.03 & \phantom{0}7.51 $\pm$ 0.51 & \phantom{0}6.13 $\pm$ 1.84 & \phantom{0}5.14 $\pm$ 0.64 & 2.56 $\pm$ 0.38 \\
\midrule
& Average rank & 6.4 & 3.5 & 6.1 & 2.8 & 3.1 & 4.8 & 1.3 \\
\bottomrule
\end{tabular}}
\vskip -0.1in
\end{table*}

\begin{table*}[h!]
\caption{Test ECE with temperature scaling (\%, $\downarrow$): Our Meta-Calibration (MC) obtains excellent calibration also after temperature scaling.}
\label{tab:main_ece_temp}
\vskip 0.1in
\centering
\resizebox{1.\linewidth}{!}{\begin{tabular}{llccccccccc}
\toprule
Dataset & Model & CE & Brier & MMCE & FL-3 & FLSD-53 & LS & MC (Ours) \\
\midrule
\multirow{4}{*}{CIFAR-10} & ResNet18 & \phantom{0}1.16 $\pm$ 0.10 (2.30) & \phantom{0}1.23 $\pm$ 0.03 (1.00) & \phantom{0}1.25 $\pm$ 0.08 (2.30) & \phantom{0}1.10 $\pm$ 0.11 (0.90) & \phantom{0}1.48 $\pm$ 0.31 (0.87) & \phantom{0}1.31 $\pm$ 0.08 (0.90) & 1.34 $\pm$ 0.47 (0.97) \\
& ResNet50 & \phantom{0}1.20 $\pm$ 0.17 (2.53) & \phantom{0}0.97 $\pm$ 0.02 (1.17) & \phantom{0}1.35 $\pm$ 0.38 (2.50) & \phantom{0}1.10 $\pm$ 0.16 (1.07) & \phantom{0}1.21 $\pm$ 0.31 (1.07) & \phantom{0}1.42 $\pm$ 0.15 (0.90) & 1.09 $\pm$ 0.09 (1.00) \\
& ResNet110 & \phantom{0}1.49 $\pm$ 0.19 (2.57) & \phantom{0}1.55 $\pm$ 0.35 (1.13) & \phantom{0}1.20 $\pm$ 0.45 (1.90) & \phantom{0}1.21 $\pm$ 0.07 (1.10) & \phantom{0}1.33 $\pm$ 0.14 (1.10) & \phantom{0}2.16 $\pm$ 0.21 (0.90) & 1.33 $\pm$ 0.37 (0.97) \\
& WideResNet26-10 & \phantom{0}1.14 $\pm$ 0.13 (2.20) & \phantom{0}1.03 $\pm$ 0.08 (1.00) & \phantom{0}0.99 $\pm$ 0.19 (2.23) & \phantom{0}1.20 $\pm$ 0.29 (0.87) & \phantom{0}1.09 $\pm$ 0.02 (0.90) & \phantom{0}1.32 $\pm$ 0.04 (0.90) & 0.94 $\pm$ 0.10 (1.00) \\
\midrule
\multirow{4}{*}{CIFAR-100} & ResNet18 & \phantom{0}5.47 $\pm$ 0.22 (1.33) & \phantom{0}4.21 $\pm$ 0.23 (0.90) & \phantom{0}6.09 $\pm$ 0.39 (1.13) & \phantom{0}2.83 $\pm$ 0.27 (1.00) & \phantom{0}2.47 $\pm$ 0.12 (1.00) & \phantom{0}4.37 $\pm$ 0.45 (0.90) & 2.71 $\pm$ 0.58 (1.03) \\
& ResNet50 & \phantom{0}2.51 $\pm$ 0.23 (1.57) & \phantom{0}3.43 $\pm$ 0.32 (1.10) & \phantom{0}3.19 $\pm$ 0.53 (1.37) & \phantom{0}2.25 $\pm$ 0.69 (1.10) & \phantom{0}2.53 $\pm$ 0.11 (1.10) & \phantom{0}4.28 $\pm$ 0.42 (1.10) & 2.51 $\pm$ 0.45 (1.07) \\
& ResNet110 & \phantom{0}3.77 $\pm$ 0.51 (1.57) & \phantom{0}3.71 $\pm$ 0.67 (1.17) & \phantom{0}2.74 $\pm$ 0.45 (1.40) & \phantom{0}3.97 $\pm$ 0.28 (1.10) & \phantom{0}4.13 $\pm$ 0.40 (1.10) & \phantom{0}6.04 $\pm$ 0.31 (1.10) & 2.55 $\pm$ 0.33 (1.03) \\
& WideResNet26-10 & \phantom{0}3.08 $\pm$ 0.26 (1.80) & \phantom{0}2.49 $\pm$ 0.13 (1.10) & \phantom{0}4.52 $\pm$ 0.52 (1.40) & \phantom{0}2.20 $\pm$ 0.12 (1.03) & \phantom{0}2.30 $\pm$ 0.36 (1.00) & \phantom{0}3.62 $\pm$ 0.74 (1.07) & 2.72 $\pm$ 0.19 (1.10) \\
\midrule
SVHN & ResNet18 & \phantom{0}0.74 $\pm$ 0.04 (2.10) & \phantom{0}0.83 $\pm$ 0.09 (0.90) & \phantom{0}1.10 $\pm$ 0.01 (2.30) &  \phantom{0}0.90 $\pm$ 0.43 (0.83) & \phantom{0}1.11 $\pm$ 0.37 (0.87) & \phantom{0}1.45 $\pm$ 0.51 (0.87) & 1.14 $\pm$ 0.12 (1.00) \\
\midrule
20 Newsgroups & Global Pooling CNN & \phantom{0}2.85 $\pm$ 0.34 (3.67) & \phantom{0}4.32 $\pm$ 0.79 (2.97) & \phantom{0}4.00 $\pm$ 0.22 (2.60) & \phantom{0}3.59 $\pm$ 0.34 (1.43) & \phantom{0}2.76 $\pm$ 0.20 (1.33) & \phantom{0}3.19 $\pm$ 0.30 (1.10) & 2.50 $\pm$ 0.32 (0.97) \\
\midrule
& Average rank & 3.7 & 3.8 & 4.4 & 3.0 & 3.9 & 6.2 & 2.8 \\
\bottomrule
\end{tabular}}
\vskip -0.1in
\end{table*}

\begin{table*}[h!]
\caption{Test error (\%, $\downarrow$): Our Meta-Calibration (MC) obtains excellent calibration with only small increases in the test error.}
\label{tab:main_error}
\vskip 0.1in
\centering
\resizebox{1.\linewidth}{!}{\begin{tabular}{llccccccccc}
\toprule
Dataset & Model & CE & Brier & MMCE & FL-3 & FLSD-53 & LS & MC (Ours) \\
\midrule
\multirow{4}{*}{CIFAR-10} & ResNet18 & \phantom{0}4.99 $\pm$ 0.14 & \phantom{0}5.27 $\pm$ 0.21 & \phantom{0}5.17 $\pm$ 0.19 & \phantom{0}5.06 $\pm$ 0.09 & \phantom{0}5.22 $\pm$ 0.04 & \phantom{0}4.94 $\pm$ 0.13 & \phantom{0}5.22 $\pm$ 0.06 \\
& ResNet50 & \phantom{0}4.90 $\pm$ 0.02 & \phantom{0}5.15 $\pm$ 0.14 & \phantom{0}5.13 $\pm$ 0.12 & \phantom{0}5.27 $\pm$ 0.22 & \phantom{0}5.26 $\pm$ 0.15 & \phantom{0}4.77 $\pm$ 0.11 & \phantom{0}5.46 $\pm$ 0.05 \\
& ResNet110 & \phantom{0}5.40 $\pm$ 0.10 & \phantom{0}5.97 $\pm$ 0.17 & \phantom{0}5.70 $\pm$ 0.12 & \phantom{0}5.67 $\pm$ 0.33 & \phantom{0}5.87 $\pm$ 0.13 & \phantom{0}5.45 $\pm$ 0.11 & \phantom{0}6.09 $\pm$ 0.22 \\
& WideResNet26-10 & \phantom{0}3.99 $\pm$ 0.07 & \phantom{0}4.20 $\pm$ 0.03 & \phantom{0}4.11 $\pm$ 0.06 & \phantom{0}4.18 $\pm$ 0.03 & \phantom{0}4.22 $\pm$ 0.05 & \phantom{0}4.05 $\pm$ 0.07 & \phantom{0}4.36 $\pm$ 0.20 \\
\midrule
\multirow{4}{*}{CIFAR-100} & ResNet18 & 22.85 $\pm$ 0.17 & 23.50 $\pm$ 0.17 & 23.80 $\pm$ 0.18 & 22.87 $\pm$ 0.16 & 23.23 $\pm$ 0.32 & 22.35 $\pm$ 0.27 & 23.88 $\pm$ 0.20 \\
& ResNet50 & 22.41 $\pm$ 0.24 & 24.81 $\pm$ 0.33 & 22.43 $\pm$ 0.05 & 22.27 $\pm$ 0.13 & 22.76 $\pm$ 0.27 & 21.85 $\pm$ 0.06 & 23.22 $\pm$ 0.48 \\
& ResNet110 & 22.99 $\pm$ 0.19 & 28.29 $\pm$ 1.42 & 23.81 $\pm$ 0.58 & 23.12 $\pm$ 0.26 & 23.71 $\pm$ 0.24 & 23.08 $\pm$ 0.15 & 24.51 $\pm$ 0.41 \\
& WideResNet26-10 & 20.41 $\pm$ 0.12 & 20.77 $\pm$ 0.05 & 20.60 $\pm$ 0.10 & 19.80 $\pm$ 0.40 & 19.97 $\pm$ 0.25 & 20.82 $\pm$ 0.42 & 22.35 $\pm$ 0.03 \\
\midrule
SVHN & ResNet18 & \phantom{0}4.11 $\pm$ 0.08 & \phantom{0}3.90 $\pm$ 0.19 & \phantom{0}4.15 $\pm$ 0.08 & \phantom{0}4.20 $\pm$ 0.07 & \phantom{0}4.18 $\pm$ 0.06 & \phantom{0}4.13 $\pm$ 0.09 & \phantom{0}4.08 $\pm$ 0.02 \\
\midrule
20 Newsgroups & Global Pooling CNN & 26.64 $\pm$ 0.27 & 26.59 $\pm$ 0.72 & 26.92 $\pm$ 0.32 & 27.65 $\pm$ 0.38 & 27.59 $\pm$ 0.94 & 26.10 $\pm$ 0.31 & 27.26 $\pm$ 0.59 \\
\midrule
& Average rank & 2.1 & 4.9 & 4.2 & 3.9 & 4.8 & 2.1 & 5.9 \\
\bottomrule
\end{tabular}}
\vskip -0.1in
\end{table*}

Note that while Brier score, Focal loss and FLSD modify the base model's loss function, our Meta-Calibration corresponds to the vanilla cross-entropy baseline, but where label smoothing is tuned by our DECE-driven hyper-parameter meta-learning rather than being selected by hand.

\subsection{Further analysis}

\paragraph{Alternative calibration metrics}
We investigate if models meta-trained using DECE also perform well when evaluated using more advanced calibration metrics than ECE. In particular, we evaluate performance using class-wise ECE (CECE) \citep{kumar2019calibration,widmann2019calibration,vaicenavicius2019calibration,kull2019calibration} that considers the scores of all classes in the predicted distribution, instead of only the class with the highest probability. The results in Table~\ref{tab:main_cece} confirm that models meta-trained using DECE have excellent calibration also in terms of the CECE criterion.

\begin{table*}[h!]
\caption{Test Classwise-ECE (\%, $\downarrow$): Our Meta-Calibration (MC) leads to excellent calibration also when using a more advanced calibration metric.}
\label{tab:main_cece}
\vskip 0.1in
\centering
\resizebox{1.\linewidth}{!}{\begin{tabular}{llccccccccc}
\toprule
Dataset & Model & CE & Brier & MMCE & FL-3 & FLSD-53 & LS-0.05 & MC (Ours) \\
\midrule
\multirow{4}{*}{CIFAR-10} & ResNet18 & \phantom{0}0.87 $\pm$ 0.04 & \phantom{0}0.46 $\pm$ 0.02 & \phantom{0}0.91 $\pm$ 0.03 & \phantom{0}0.52 $\pm$ 0.02 & \phantom{0}0.53 $\pm$ 0.03 & \phantom{0}0.73 $\pm$ 0.01 & \phantom{0}0.41 $\pm$ 0.01 \\
& ResNet50 & \phantom{0}0.88 $\pm$ 0.01 & \phantom{0}0.46 $\pm$ 0.02 & \phantom{0}0.93 $\pm$ 0.04 & \phantom{0}0.44 $\pm$ 0.02 & \phantom{0}0.44 $\pm$ 0.03 & \phantom{0}0.63 $\pm$ 0.02 & \phantom{0}0.48 $\pm$ 0.04 \\
& ResNet110 & \phantom{0}0.99 $\pm$ 0.02 & \phantom{0}0.58 $\pm$ 0.04 & \phantom{0}0.89 $\pm$ 0.14 & \phantom{0}0.50 $\pm$ 0.02 & \phantom{0}0.55 $\pm$ 0.05 & \phantom{0}0.66 $\pm$ 0.03 & \phantom{0}0.45 $\pm$ 0.01 \\
& WideResNet & \phantom{0}0.71 $\pm$ 0.01 & \phantom{0}0.37 $\pm$ 0.00 & \phantom{0}0.74 $\pm$ 0.01 & \phantom{0}0.43 $\pm$ 0.02 & \phantom{0}0.43 $\pm$ 0.04 & \phantom{0}0.72 $\pm$ 0.02 & \phantom{0}0.34 $\pm$ 0.00 \\
\midrule
\multirow{4}{*}{CIFAR-100} & ResNet18 & \phantom{0}0.23 $\pm$ 0.01 & \phantom{0}0.24 $\pm$ 0.00 & \phantom{0}0.22 $\pm$ 0.01 & \phantom{0}0.20 $\pm$ 0.00 & \phantom{0}0.20 $\pm$ 0.00 & \phantom{0}0.26 $\pm$ 0.00 & \phantom{0}0.19 $\pm$ 0.00 \\
& ResNet50 & \phantom{0}0.29 $\pm$ 0.03 & \phantom{0}0.20 $\pm$ 0.01 & \phantom{0}0.24 $\pm$ 0.03 & \phantom{0}0.20 $\pm$ 0.00 & \phantom{0}0.20 $\pm$ 0.01 & \phantom{0}0.21 $\pm$ 0.00 & \phantom{0}0.19 $\pm$ 0.00 \\
& ResNet110 & \phantom{0}0.34 $\pm$ 0.02 & \phantom{0}0.23 $\pm$ 0.01 & \phantom{0}0.29 $\pm$ 0.02 & \phantom{0}0.22 $\pm$ 0.01 & \phantom{0}0.23 $\pm$ 0.00 & \phantom{0}0.26 $\pm$ 0.00 & \phantom{0}0.19 $\pm$ 0.00 \\
& WideResNet & \phantom{0}0.29 $\pm$ 0.02 & \phantom{0}0.19 $\pm$ 0.00 & \phantom{0}0.23 $\pm$ 0.03 & \phantom{0}0.18 $\pm$ 0.00 & \phantom{0}0.18 $\pm$ 0.00 & \phantom{0}0.21 $\pm$ 0.01 & \phantom{0}0.19 $\pm$ 0.00 \\
\midrule
SVHN & ResNet18 & \phantom{0}0.62 $\pm$ 0.02 & \phantom{0}0.52 $\pm$ 0.02 & \phantom{0}0.65 $\pm$ 0.02 & \phantom{0}0.67 $\pm$ 0.03 & \phantom{0}0.68 $\pm$ 0.04 & \phantom{0}0.81 $\pm$ 0.05 & \phantom{0}0.29 $\pm$ 0.03 \\
\midrule
20 Newsgroups & Global Pooling CNN & \phantom{0}2.01 $\pm$ 0.07 & \phantom{0}1.80 $\pm$ 0.04 & \phantom{0}1.63 $\pm$ 0.09 & \phantom{0}1.29 $\pm$ 0.03 & \phantom{0}1.22 $\pm$ 0.14 & \phantom{0}0.97 $\pm$ 0.06 & \phantom{0}1.00 $\pm$ 0.04 \\
\midrule
& Average rank & 6.0 & 3.3 & 5.8 & 2.5 & 2.8 & 5.1 & 1.6 \\
\bottomrule
\end{tabular}}
\vskip -0.1in
\end{table*}

\paragraph{Alternative hyper-parameter choice}
We present a general metric that can be used for optimising hyper-parameters for superior calibration. While our main experiments are conducted with non-uniform label smoothing, we demonstrate the generality of the framework by also learning alternative meta-parameters. In particular, we also consider scalar and vector version of label smoothing as well as learnable L2 regularization. We perform the additional evaluation using ResNet18 on the CIFAR benchmark.

The results in Table \ref{tab:l2comparison} confirm learnable L2 regularization also leads to clear improvement in ECE over the cross-entropy baseline. However, the error rate is slightly increased compared to learnable LS, hence we focused on the latter for our other experiments. Scalar and vector LS (MC-S and MC-V) have both improved the calibration, but non-uniform label smoothing (MC) has worked better thanks to its larger expressivity.

\begin{table}[h!]
\caption{Comparison of hyper-parameter choice for meta-calibration: CIFAR benchmark with ResNet18 model. Test errors (\%, $\downarrow$) and test ECE (\%, $\downarrow$). Other variants of Meta-Calibration also lead to strong improvements in calibration, with non-uniform label smoothing leading to the best calibration overall.}
\vskip 0.1in
\label{tab:l2comparison}
\centering
\resizebox{0.45\linewidth}{!}{\begin{tabular}{llcccc}
\toprule
Dataset & Method & ECE ($\downarrow$) & Error ($\downarrow$) \\
\midrule
\multirow{5}{*}{CIFAR-10} & CE & 4.23 $\pm$ 0.15 & \phantom{0}4.99 $\pm$ 0.14 \\
& MC & 1.17 $\pm$ 0.26 & \phantom{0}5.22 $\pm$ 0.06 \\
& MC-S & 1.48 $\pm$ 0.26 & \phantom{0}5.17 $\pm$ 0.13 \\
& MC-V & 1.51 $\pm$ 0.26 & \phantom{0}5.07 $\pm$ 0.03 \\
& MC-L2 & 1.78 $\pm$ 0.22 & \phantom{0}5.49 $\pm$ 0.14 \\
\midrule
\multirow{5}{*}{CIFAR-100} & CE & 8.79 $\pm$ 0.59 & 22.85 $\pm$ 0.17 \\
& MC & 2.52 $\pm$ 0.35 & 23.88 $\pm$ 0.20 \\
& MC-S & 6.13 $\pm$ 1.20 & 24.07 $\pm$ 0.17 \\
& MC-V & 3.98 $\pm$ 0.23 & 23.96 $\pm$ 0.12 \\
& MC-L2 & 4.18 $\pm$ 0.26 & 26.10 $\pm$ 0.14 \\
\bottomrule
\end{tabular}}
\vskip -0.1in
\end{table}

\paragraph{Ablation study on meta-learning objective design}
Recall our framework in Equation~\ref{eq:BLO} is setup to perform conventional model training in the inner optimisation, given hyper-parameters; and meta-learning of hyper-parameters in the outer optimisation, by minimising a combination of cross-entropy and our DECE metric as evaluated on the meta-validation set. While we view this setup as being the most intuitive, other architectures are also possible in terms of choice of objective for use in the inner and outer layer of the bilevel optimisation. As a comparison to our DECE, we also evaluate the prior metric MMCE previously proposed as a proxy for model calibration in \citep{Kumar2018TrainableEmbeddings}.

From the results in Table \ref{tab:ablationCompact} we can conclude that: 1) Meta-learning with combined CE and DECE meta-objective is beneficial for improving calibration (M5 vs M0). 2) Alternative outer-loop objectives CE (M2) and DECE (M3) improve calibration but not as significantly as the combined meta-objective (M5 vs M2 and M5 vs M4). 3) MMCE completely fails as a meta-objective (M3). 4) DECE improves calibration when used as a secondary loss in multi-task learning, but at greater detriment to test error (M1 vs M0). 5) Our combined meta-objective (M5) is the best overall.

\begin{table*}[h!]
\caption{Ablation study on losses for inner and outer objectives in bilevel optimisation using CIFAR-10 and CIFAR-100 with ResNet18.}
\label{tab:ablationCompact}
\vskip 0.1in
\centering
\resizebox{\textwidth}{!}{
\begin{tabular}{lccccccc}
\toprule
 &  &  & \multicolumn{2}{c}{CIFAR-10} & \multicolumn{2}{c}{CIFAR-100} \\
Model & Meta-Loss & Loss & ECE (\%, $\downarrow$) & Error (\%, $\downarrow$) & ECE (\%, $\downarrow$) & Error (\%, $\downarrow$) \\
\midrule
M0: Vanilla CE & - & CE & \phantom{0}4.23 $\pm$ 0.15 & \phantom{0}4.99 $\pm$ 0.14 & \phantom{0}8.79 $\pm$ 0.59 & 22.85 $\pm$ 0.17 \\
M1: Multi-task & - & CE + DECE & \phantom{0}3.80 $\pm$ 0.03 & 10.24 $\pm$ 0.21 & \phantom{0}4.40 $\pm$ 0.39 & 29.49 $\pm$ 0.17 \\
M2: Meta-Calibration & CE & CE & \phantom{0}1.31 $\pm$ 0.36 & \phantom{0}5.13 $\pm$ 0.24 & \phantom{0}3.00 $\pm$ 1.12 & 23.72 $\pm$ 0.40 \\
M3: Meta-Calibration & MMCE & CE & 44.24 $\pm$ 0.70 & \phantom{0}6.77 $\pm$ 0.25 & 21.94 $\pm$ 2.39 & 25.40 $\pm$ 0.31 \\
M4: Meta-Calibration & DECE & CE & \phantom{0}1.26 $\pm$ 0.44 & \phantom{0}5.21 $\pm$ 0.14 & \phantom{0}3.28 $\pm$ 0.31 & 23.83 $\pm$ 0.14 \\
M5: Meta-Calibration & CE + DECE & CE & \phantom{0}1.17 $\pm$ 0.26 & \phantom{0}5.22 $\pm$ 0.06 & \phantom{0}2.52 $\pm$ 0.35 & 23.88 $\pm$ 0.20 \\
\bottomrule
\end{tabular}}
\vskip -0.1in
\end{table*}

\begin{figure}[h!]
    \centering
    \begin{subfigure}[b]{0.48\linewidth}        %% or \columnwidth
        \centering
        \includegraphics[width=\linewidth]{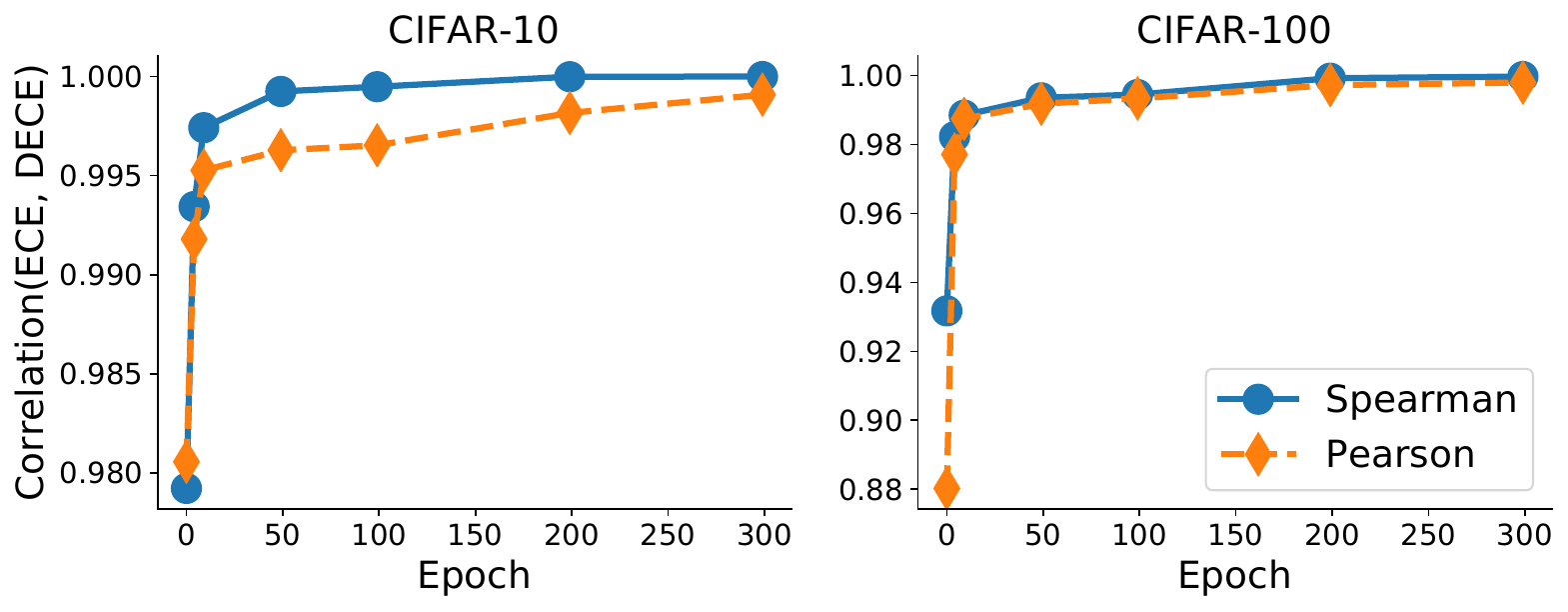}
        \caption{Correlation between DECE and ECE is close to 1.}
        \label{fig:corr-analysis}
    \end{subfigure}
    \begin{subfigure}[b]{0.48\linewidth}        %% or \columnwidth
        \centering
        \includegraphics[width=\linewidth]{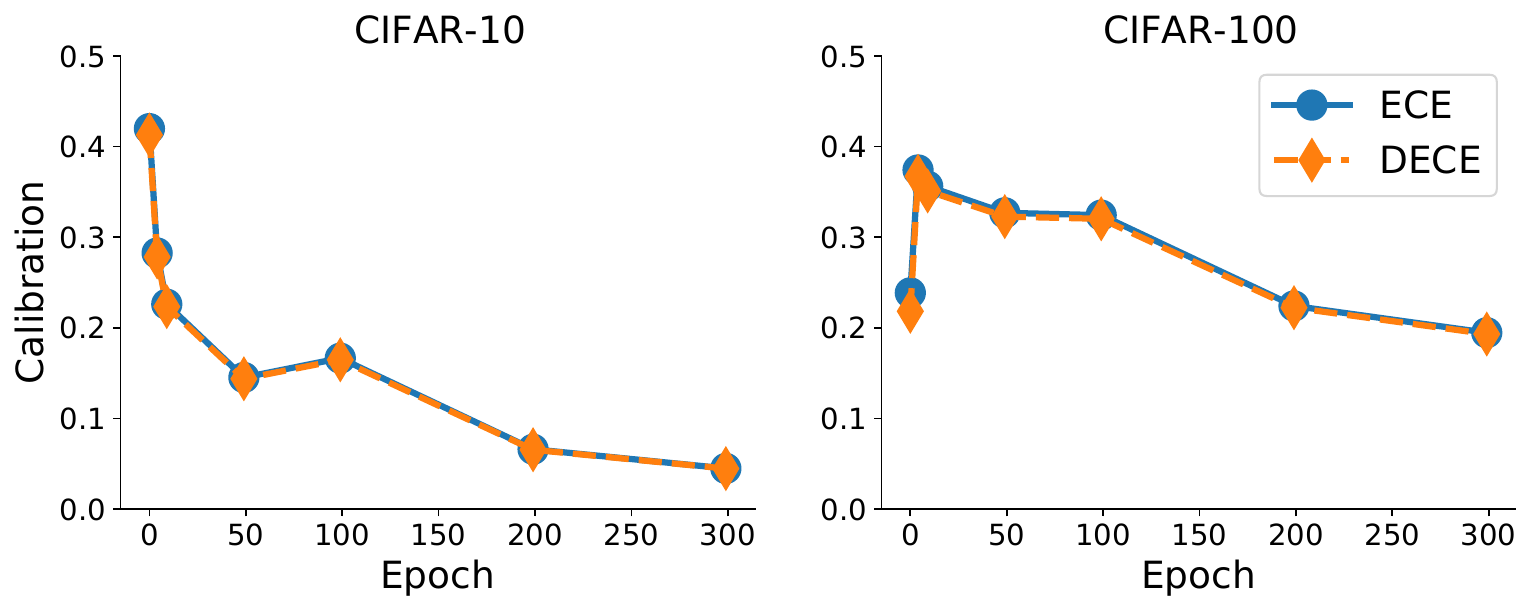}
        \caption{Mean ECE and DECE are close to each other.}
        \label{fig:value-ece-analysis}
    \end{subfigure}
    \caption{Evaluation of how DECE approximates ECE, using ResNet18 on CIFAR.}
    \label{fig:ece-dece-analysis}
\end{figure}

\paragraph{Evaluating DECE approximation to ECE}
A key contribution of this work is DECE, a differentiable approximation to expected calibration error. In this section we investigate the quality of our DECE approximation.  We trained the same ResNet18 backbone on both CIFAR-10 and CIFAR-100 benchmarks for 350 epochs, recording DECE and ECE values at various points.
The results in Figure \ref{fig:corr-analysis} show both Spearman and Pearson correlation coefficient between DECE and ECE. In both cases they are close to 1, and become even closer to 1 as training continues. This shows that DECE accurately estimates ECE, while providing differentiability for end-to-end optimisation. We further we show in Figure \ref{fig:value-ece-analysis} that their mean values are very close to each other.

\begin{figure}[h!]
\vskip 0.1in
\begin{center}
\centerline{\includegraphics[width=0.8\columnwidth]{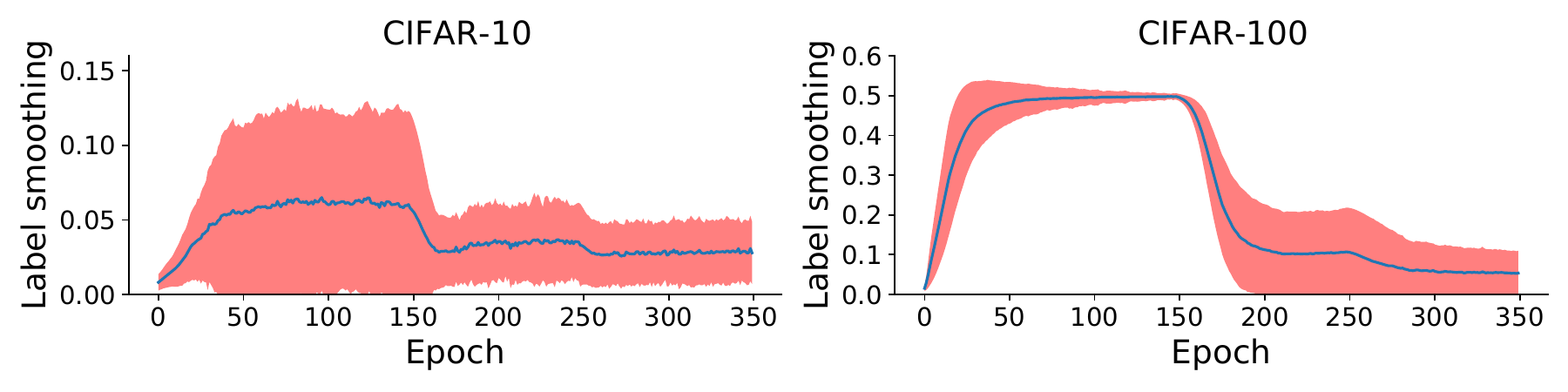}}
\caption{Overall label smoothing during training for CIFAR, using ResNet18. The learned smoothing strategy is non-trivial and adapts according to the learning rate schedule.}
\label{fig:label-analysis}
\end{center}
\vskip -0.1in
\end{figure}

\paragraph{What hyper-parameters are learned?}
We show our approach learns non-trivial hyper-parameter settings to achieve its excellent calibration performance. Figure \ref{fig:label-analysis} shows how the learned overall strength of smoothing evolves during training for both CIFAR-10 and CIFAR-100 benchmarks -- using ResNet18. We show the mean and standard deviation across three repetitions and all classes.

From the figure we observe label smoothing changes in response to changes in learning rate, which happens after 150 and 250 epochs. For CIFAR-100 with more classes it starts with large smoothing values and finishes with smaller values. The large standard deviations are due to the model making use of a wide range of class-wise smoothing parameters. It would be infeasible to manually select a curriculum for label smoothing at different stages of training, as it would be to tune a range of smoothing parameters: The ability to optimise these hyper-parameters automatically is a key benefit of our framework.  

\begin{figure}[h!]
    \centering
    \begin{subfigure}[b]{0.48\linewidth}        %% or \columnwidth
        \centering
        \includegraphics[width=\linewidth]{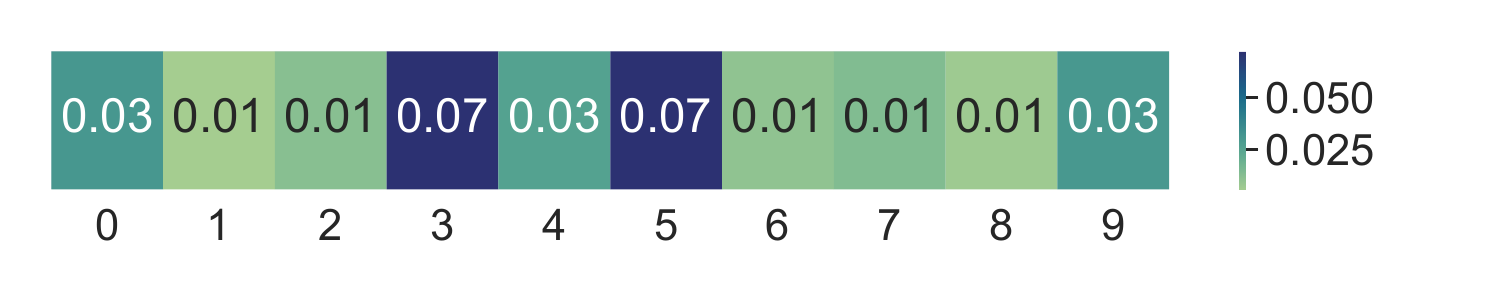}
        \vphantom{\includegraphics[height=0.2\linewidth]{example-image-10x16}}
        \caption{Overall level of smoothing in various classes.}
        \label{fig:overall-smoothing-c10}
    \end{subfigure}
    \begin{subfigure}[b]{0.48\linewidth}        %% or \columnwidth
        \centering
        \includegraphics[width=\linewidth]{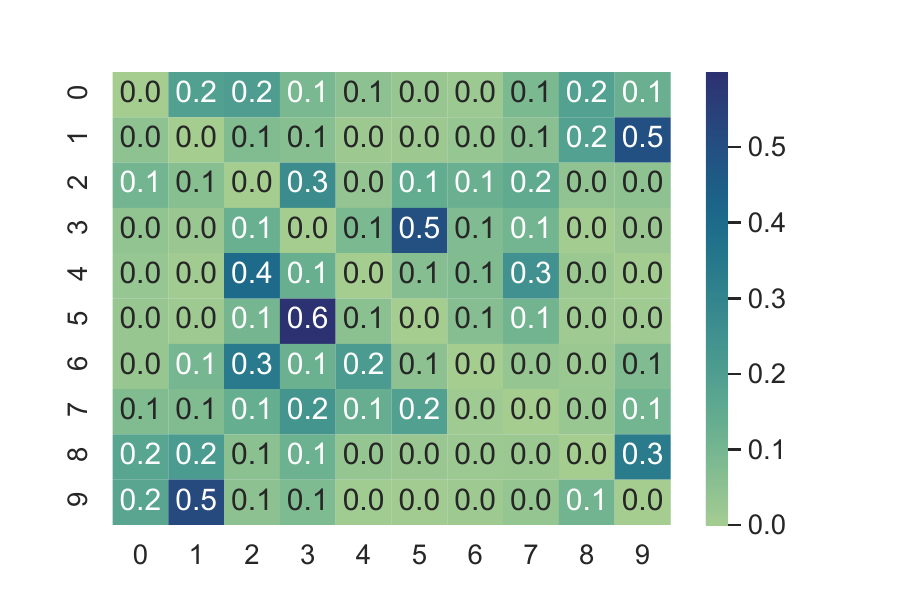}
        \caption{Distribution of smoothing across various classes.}
        \label{fig:distribution-smoothing-c10}
    \end{subfigure}
    \caption{Analysis of learned non-uniform label smoothing for CIFAR-10, using ResNet18 model. Visually similar classes receive more smoothing -- e.g. cat and dog (3 and 5), and automobile and truck (1 and 9).}
    \label{fig:dist-smoothing-analysis}
\end{figure}

\begin{figure}[h!]
\vskip 0.1in
\begin{center}
\centerline{\includegraphics[width=1.0\columnwidth]{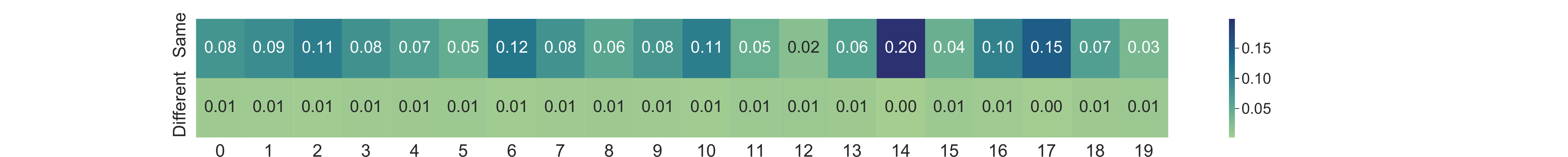}}
\caption{Meta-Calibration learns to give more label smoothing within the same superclass compared to other superclasses in CIFAR-100, using ResNet18 model.}
\label{fig:superclass-smoothing}
\end{center}
\vskip -0.1in
\end{figure}

We also analyse how the smoothing is distributed across the different classes in Figure \ref{fig:dist-smoothing-analysis} and \ref{fig:superclass-smoothing}. The results show that the smoothing is indeed non-uniform, demonstrating the model does exploit the ability to learn a complex label-smoothing distribution. The learned non-uniform label-smoothing distribution can be observed to subject visually similar classes to more smoothing (Figure~\ref{fig:dist-smoothing-analysis}(b)), which makes sense to reduce the confidence of the most likely kinds of specific errors. This idea is quantified more systematically for CIFAR-100 in Figure~\ref{fig:superclass-smoothing}, which compares the average degree of smoothing between classes in the same superclass, and those in different superclasses. The results show that within-superclass smoothing is generally much stronger than across-superclass smoothing, even though the model receives no annotation or supervision about superclasses. It learns this smoothing structure given the objective of optimising (meta-)validation calibration.

We further analysed the hyper-parameters in the case of learnable L2 regularization and show it as part of the appendix. The figure shows we learn a range of regularization values to achieve a good calibration outcome. This highlights the value of our differentiable framework that enables efficient gradient-based optimisation of many hyper-parameters.

\paragraph{Analysis of accuracy vs calibration using simulated data}
Meta-Calibration leads to significantly better calibration, while keeping similar or only marginally worse accuracy. To study if there is a trade-off between accuracy vs calibration we perform an experiment using simulated data. More specifically the experiment with simulated data consists of 1) sampling parameters of a binary logistic regression model (oracle) over 2D data, 2) sampling the label for each data point from a binomial distribution with a class probability given by the oracle. 3) We then use the sampled data to train (i) a vanilla logistic regression model, (ii) label-smoothing logistic regression model and (iii) a Meta-Calibration logistic regression model.  This enables us to compare the best-case classifier accuracy and calibration with the results of learned models. Our experiment is repeated across three random seeds so that we can report the mean and standard deviation.

\begin{figure}[t]
\vskip 0.1in
\begin{center}
\centerline{\includegraphics[width=0.75\columnwidth]{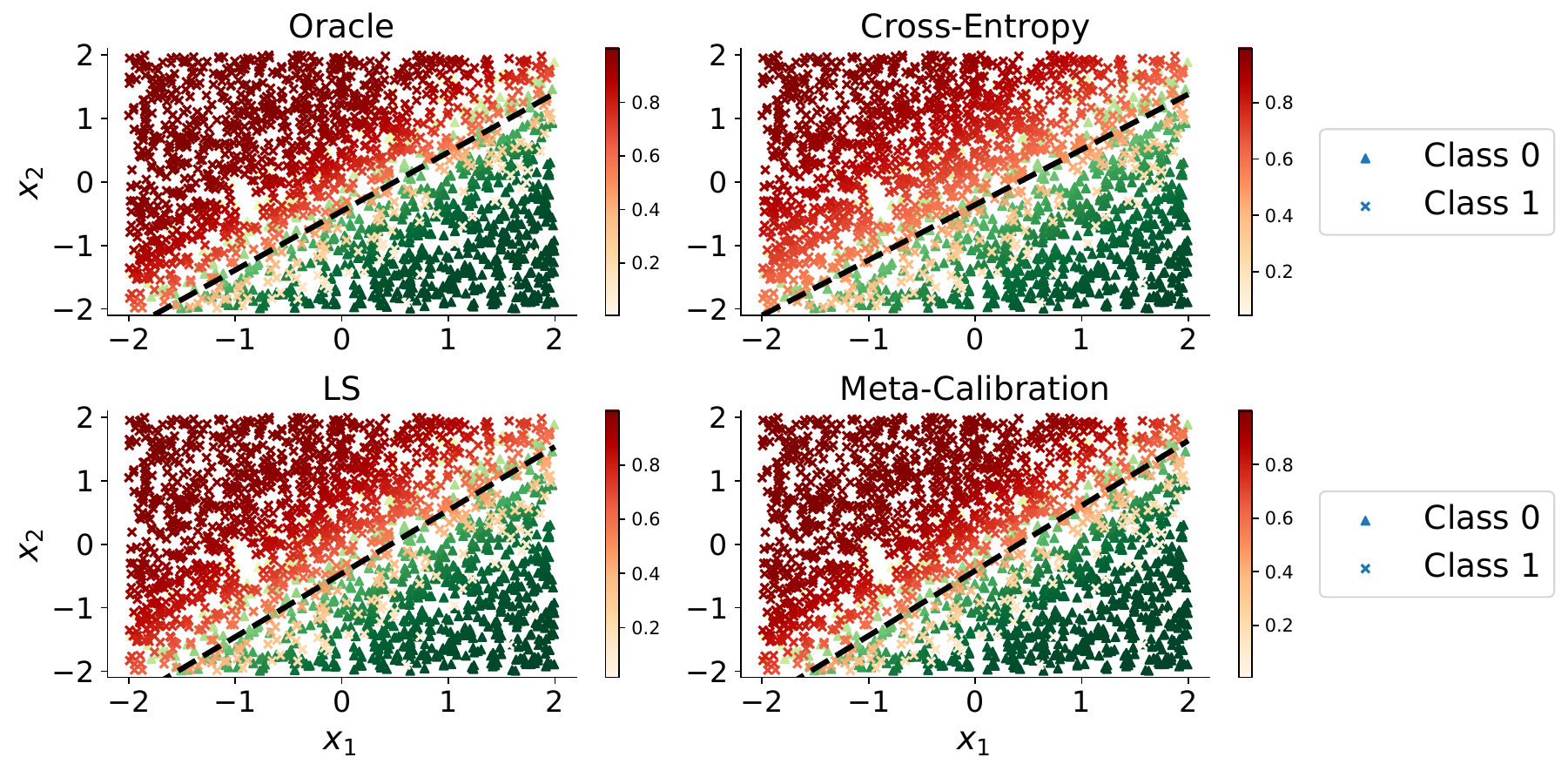}}
\caption{Visualization of the estimated probabilities across test data points, with the decision boundary.}
\label{fig:synthetic}
\end{center}
\vskip -0.1in
\end{figure}

The results in Table \ref{tab:synthetic} confirm Meta-Calibration matches both the test accuracy and ECE of the oracle, obtaining close to the best possible calibration. Analogous cross-entropy training as well as simple label smoothing (LS) have a significantly larger ECE. The ECE of Meta-Calibration and oracle is close to 0, but not precisely 0 due to sampling effects (i.e. because we do not use an infinite amount of data). We have also visualized the estimated class probabilities of different data points, together with the decision boundary in Figure~\ref{fig:synthetic}. The visualization shows that Meta-Calibration and the oracle are similarly calibrated (e.g., similar point shades close to the decision boundary), while a difference in calibration (point shading) is perceptible for cross-entropy and LS.

\begin{table*}[h!]
\caption{Analysis of accuracy and calibration on simulated data with known oracle. Test accuracy and ECE (\%) are reported, with the mean and standard deviation computed across three random seeds. Our Meta-Calibration (MC) matches both the accuracy and ECE of the oracle, obtaining close to perfect calibration.}
\label{tab:synthetic}
\vskip 0.1in
\centering
\begin{tabular}{lcccc}
\toprule
Metric & Oracle & Cross-Entropy & LS & MC (Ours) \\
\midrule
Accuracy ($\uparrow$) & 87.62 $\pm$ 1.87 & 87.55 $\pm$ 1.86 & 87.57 $\pm$ 1.87 & 87.55 $\pm$ 1.81 \\
ECE ($\downarrow$) & \phantom{0}1.38 $\pm$ 0.32 & \phantom{0}9.81 $\pm$ 1.14 & \phantom{0}3.61 $\pm$ 0.43 & \phantom{0}1.40 $\pm$ 0.19 \\
\bottomrule
\end{tabular}
\vskip -0.1in
\end{table*}

\section{Discussion}
This work is a pioneering step in using meta-learning to directly optimise model calibration. Learnable rather than hand-tuned calibration is important as different models and datasets have very different calibration properties, precluding a one-size-fits-all solution \citep{minderer2021revisiting}. There are many ways our work could be extended in the future. One direction is to target different hyper-parameters beyond the label smoothing and L2 classifier regularization evaluated here, such as loss learning. While we did not explore this here, the framework could also be used to unify post-hoc methods such as temperature scaling by treating temperature as the hyper-parameter. Secondly, better meta-learning algorithms such as implicit meta-learning \citep{Lorraine2020optimizingDifferentiation} could better optimise the meta-objective. A third direction is to extend the differentiable metric itself to e.g. adapt it to various domain-specific calibration measures -- for example ones relevant to the finance industry \citep{Liu2019ACalibration}.

A drawback of our approach is the computational overhead added by meta-learning compared to basic model training. However, it is still manageable and may be worth it when well-calibrated models are crucial. We give an overview in Table~\ref{tab:time} in the appendix.
Ongoing advances in efficient meta-learners  \citep{Bohdal2021EvoGrad:optimization} can make the overhead smaller. In terms of social implications, our work aims to improve the reliability of neural networks, but there still are risks the neural networks will fail to accurately estimate their confidence.

\section{Conclusion}
We introduced a new DECE metric that accurately represents the common calibration ECE measure and makes it differentiable. With DECE, we can directly optimise hyper-parameters for calibration and obtain competitive results with hand-designed architectures. We believe DECE opens up a new avenue for the community to tackle the challenge of model calibration in optimisation-based ways.

\subsubsection*{Acknowledgments}
This work was supported in part by the EPSRC Centre for Doctoral Training in Data Science, funded by the UK Engineering and Physical Sciences Research Council (grant EP/L016427/1) and the University of Edinburgh.

\bibliographystyle{tmlr}
\bibliography{references,morerefs}

\begin{thebibliography}{64}
\providecommand{\natexlab}[1]{#1}
\providecommand{\url}[1]{\texttt{#1}}
\expandafter\ifx\csname urlstyle\endcsname\relax
  \providecommand{\doi}[1]{doi: #1}\else
  \providecommand{\doi}{doi: \begingroup \urlstyle{rm}\Url}\fi

\bibitem[Algan \& Ulusoy(2022)Algan and Ulusoy]{algan2022noisy}
Gorkem Algan and Ilkay Ulusoy.
\newblock Metalabelnet: Learning to generate soft-labels from noisy-labels.
\newblock \emph{IEEE Transactions on Image Processing}, 2022.

\bibitem[Balaji et~al.(2018)Balaji, Sankaranarayanan, and
  Chellappa]{Balaji2018MetaReg:Meta-regularization}
Yogesh Balaji, Swami Sankaranarayanan, and Rama Chellappa.
\newblock {MetaReg: towards domain generalization using meta-regularization}.
\newblock In \emph{NeurIPS}, 2018.

\bibitem[Blondel et~al.(2020)Blondel, Teboul, Berthet, and
  Djolonga]{Blondel2020FastRanking}
Mathieu Blondel, Olivier Teboul, Quentin Berthet, and Josip Djolonga.
\newblock {Fast differentiable sorting and ranking}.
\newblock In \emph{ICML}, 2020.

\bibitem[Bohdal et~al.(2020)Bohdal, Yang, and
  Hospedales]{Bohdal2020FlexibleImages}
Ondrej Bohdal, Yongxin Yang, and Timothy Hospedales.
\newblock {Flexible dataset distillation: learn labels instead of images}.
\newblock In \emph{NeurIPS MetaLearn Workshop}, 2020.

\bibitem[Bohdal et~al.(2021)Bohdal, Yang, and
  Hospedales]{Bohdal2021EvoGrad:optimization}
Ondrej Bohdal, Yongxin Yang, and Timothy Hospedales.
\newblock {EvoGrad: efficient gradient-based meta-learning and hyperparameter
  optimization}.
\newblock In \emph{NeurIPS}, 2021.

\bibitem[Bohdal et~al.(2022)Bohdal, Li, Hu, and
  Hospedales]{Bohdal2022Feed-forwardCross-attention}
Ondrej Bohdal, Da~Li, Shell~Xu Hu, and Timothy Hospedales.
\newblock {Feed-forward source-free latent domain adaptation via
  cross-attention}.
\newblock In \emph{ICML 2022 Workshop on Pre-training: Perspectives, Pitfalls,
  and Paths Forward}, 2022.

\bibitem[Bohdal et~al.(2023)Bohdal, Tian, Zong, Chavhan, Li, Gouk, Guo, and
  Hospedales]{Bohdal2023MetaLearning-to-learn}
Ondrej Bohdal, Yinbing Tian, Yongshuo Zong, Ruchika Chavhan, Da~Li, Henry Gouk,
  Li~Guo, and Timothy Hospedales.
\newblock {Meta Omnium: A benchmark for general-purpose learning-to-learn}.
\newblock In \emph{CVPR}, 2023.

\bibitem[Bojarski et~al.(2016)Bojarski, Del~Testa, Dworakowski, Firner, Flepp,
  Goyal, Jackel, Monfort, Muller, Zhang, Zhang, Zhao, and
  Zieba]{Bojarski2016EndCars}
Mariusz Bojarski, Davide Del~Testa, Daniel Dworakowski, Bernhard Firner, Beat
  Flepp, Prasoon Goyal, Lawrence~D Jackel, Mathew Monfort, Urs Muller, Jiakai
  Zhang, Xin Zhang, Jake Zhao, and Karol Zieba.
\newblock {End to end learning for self-driving cars}.
\newblock In \emph{arXiv}, 2016.

\bibitem[Brier(1950)]{Brier1950VerificationProbability}
Glenn~W Brier.
\newblock {Verification of forecasts expressed in terms of probability}.
\newblock In \emph{Monthly weather review}, 1950.

\bibitem[Carrell et~al.(2022)Carrell, Mallinar, Lucas, and
  Nakkiran]{carrell2022calibration}
Annabelle Carrell, Neil Mallinar, James Lucas, and Preetum Nakkiran.
\newblock The calibration generalization gap.
\newblock In \emph{ICML 2022 Workshop on Distribution-Free Uncertainty
  Quantification}, 2022.

\bibitem[Caruana et~al.(2015)Caruana, Lou, Gehrke, Koch, Sturm, and
  Elhadad]{Caruana2015IntelligibleHealthcare}
Rich Caruana, Yin Lou, Johannes Gehrke, Paul Koch, Marc Sturm, and Noemie
  Elhadad.
\newblock {Intelligible models for healthcare}.
\newblock In \emph{KDD}, 2015.

\bibitem[Chen et~al.(2023)Chen, Tripp, and
  Hern{\'a}ndez-Lobato]{Chen2022MetalearningAD}
Wenlin Chen, Austin Tripp, and Jos{\'e}~Miguel Hern{\'a}ndez-Lobato.
\newblock Meta-learning adaptive deep kernel gaussian processes for molecular
  property prediction.
\newblock In \emph{ICLR}, 2023.

\bibitem[El-Sappagh et~al.(2023)El-Sappagh, Alonso-Moral, Abuhmed, Ali, and
  Bugar{\'i}n-Diz]{ElSappagh2023TrustworthyAI}
Shaker El-Sappagh, Jose~Maria Alonso-Moral, Tamer Abuhmed, Farman Ali, and
  Alberto Bugar{\'i}n-Diz.
\newblock Trustworthy artificial intelligence in alzheimer’s disease: state
  of the art, opportunities, and challenges.
\newblock \emph{Artificial Intelligence Review}, 2023.

\bibitem[Ferianc et~al.(2023)Ferianc, Bohdal, Hospedales, and
  Rodrigues]{Ferianc2023ImpactNetworks}
Martin Ferianc, Ondrej Bohdal, Timothy Hospedales, and Miguel Rodrigues.
\newblock {Impact of noise on calibration and generalisation of neural
  networks}.
\newblock In \emph{ICML 2023 Workshop on Spurious Correlations, Invariance, and
  Stability}, 2023.

\bibitem[Finn et~al.(2017)Finn, Abbeel, and
  Levine]{Finn2017Model-agnosticNetworks}
Chelsea Finn, Pieter Abbeel, and Sergey Levine.
\newblock {Model-agnostic meta-learning for fast adaptation of deep networks}.
\newblock In \emph{ICML}, 2017.

\bibitem[Gawlikowski et~al.(2021)Gawlikowski, Tassi, Ali, Lee, Humt, Feng,
  Kruspe, Triebel, Jung, Roscher, Shahzad, Yang, Bamler, and
  Zhu]{Gawlikowski2021ASO}
Jakob Gawlikowski, Cedrique Rovile~Njieutcheu Tassi, Mohsin Ali, Jongseo Lee,
  Matthias Humt, Jianxiang Feng, Anna~M. Kruspe, Rudolph Triebel, Peter Jung,
  Ribana Roscher, M.~Shahzad, Wen Yang, Richard Bamler, and Xiaoxiang Zhu.
\newblock A survey of uncertainty in deep neural networks.
\newblock \emph{ArXiv}, 2021.

\bibitem[Guo et~al.(2017)Guo, Pleiss, Sun, and Weinberger]{Guo2017OnNetworks}
Chuan Guo, Geoff Pleiss, Yu~Sun, and Kilian~Q Weinberger.
\newblock {On calibration of modern neural networks}.
\newblock In \emph{ICML}, 2017.

\bibitem[He et~al.(2015)He, Zhang, Ren, and Sun]{He2015DeepRecognition}
Kaiming He, Xiangyu Zhang, Shaoqing Ren, and Jian Sun.
\newblock {Deep residual learning for image recognition}.
\newblock In \emph{CVPR}, 2015.

\bibitem[Hendrycks \& Dietterich(2019)Hendrycks and
  Dietterich]{Hendrycks2019BenchmarkingPerturbations}
Dan Hendrycks and Thomas Dietterich.
\newblock {Benchmarking neural network robustness to common corruptions and
  perturbations}.
\newblock In \emph{ICLR}, 2019.

\bibitem[Hendrycks \& Gimpel(2017)Hendrycks and Gimpel]{Hendrycks2017ANetworks}
Dan Hendrycks and Kevin Gimpel.
\newblock {A baseline for detecting misclassified and out-of-distribution
  examples in neural networks}.
\newblock In \emph{ICLR}, 2017.

\bibitem[Hospedales et~al.(2021)Hospedales, Antoniou, Micaelli, and
  Storkey]{Hospedales2021Meta-learningSurvey}
Timothy~M Hospedales, Antreas Antoniou, Paul Micaelli, and Amos~J Storkey.
\newblock {Meta-learning in neural networks: a survey}.
\newblock \emph{Transactions on Pattern Analysis and Machine Intelligence},
  2021.

\bibitem[Jiang et~al.(2012)Jiang, Osl, Kim, and
  Ohno-Machado]{Jiang2012CalibratingMedicine}
Xiaoqian Jiang, Melanie Osl, Jihoon Kim, and Lucila Ohno-Machado.
\newblock {Calibrating predictive model estimates to support personalized
  medicine}.
\newblock \emph{Journal of the American Medical Informatics Association}, 2012.

\bibitem[Judge et~al.(2022)Judge, Bernard, Porumb, Chartsias, Beqiri, and
  Jodoin]{Judge2022CRISPR}
Thierry Judge, Olivier Bernard, Mihaela Porumb, Agisilaos Chartsias, Arian
  Beqiri, and Pierre-Marc Jodoin.
\newblock Crisp - reliable uncertainty estimation for medical image
  segmentation.
\newblock In \emph{MICCAI}, 2022.

\bibitem[Karandikar et~al.(2021)Karandikar, Cain, Tran, Lakshminarayanan,
  Shlens, Mozer, and Roelofs]{Karandikar2021SoftNetworks}
Archit Karandikar, Nicholas Cain, Dustin Tran, Balaji Lakshminarayanan,
  Jonathon Shlens, Michael~C Mozer, and Becca Roelofs.
\newblock {Soft calibration objectives for neural networks}.
\newblock In \emph{NeurIPS}, 2021.

\bibitem[Kim \& Yun(2020)Kim and Yun]{Kim2020TaskClassification}
Sungnyun Kim and Se-Young Yun.
\newblock {Task calibration for distributional uncertainty in few-shot
  classification}.
\newblock In \emph{OpenReview}, 2020.

\bibitem[Kingma \& Ba(2015)Kingma and Ba]{Kingma2015Adam:optimization}
Diederik~P Kingma and Jimmy Ba.
\newblock {Adam: a method for stochastic optimization}.
\newblock In \emph{ICLR}, 2015.

\bibitem[Krishnan \& Tickoo(2020)Krishnan and Tickoo]{krishnan2020improving}
Ranganath Krishnan and Omesh Tickoo.
\newblock Improving model calibration with accuracy versus uncertainty
  optimization.
\newblock In \emph{NeurIPS}, 2020.

\bibitem[Krizhevsky(2009)]{Krizhevsky2009LearningImages}
Alex Krizhevsky.
\newblock {Learning multiple layers of features from tiny images}.
\newblock Technical report, University of Toronto, 2009.

\bibitem[Kull et~al.(2019)Kull, Perello~Nieto, K\"{a}ngsepp, Silva~Filho, Song,
  and Flach]{kull2019calibration}
Meelis Kull, Miquel Perello~Nieto, Markus K\"{a}ngsepp, Telmo Silva~Filho, Hao
  Song, and Peter Flach.
\newblock Beyond temperature scaling: Obtaining well-calibrated multi-class
  probabilities with dirichlet calibration.
\newblock In \emph{NeurIPS}, 2019.

\bibitem[Kumar et~al.(2019)Kumar, Liang, and Ma]{kumar2019calibration}
Ananya Kumar, Percy Liang, and Tengyu Ma.
\newblock Verified uncertainty calibration.
\newblock In \emph{NeurIPS}, 2019.

\bibitem[Kumar et~al.(2018)Kumar, Sarawagi, and
  Jain]{Kumar2018TrainableEmbeddings}
Aviral Kumar, Sunita Sarawagi, and Ujjwal Jain.
\newblock {Trainable calibration measures for neural networks from kernel mean
  embeddings}.
\newblock In \emph{ICML}, 2018.

\bibitem[Lang(1995)]{Lang1995Newsweeder:Netnews}
Ken Lang.
\newblock {Newsweeder: Learning to filter netnews}.
\newblock In \emph{ICML}, 1995.

\bibitem[Li et~al.(2020)Li, Foo, and Bilen]{Li2020LearningLearning}
Wei-Hong Li, Chuan-Sheng Foo, and Hakan Bilen.
\newblock {Learning to impute: a general framework for semi-supervised
  learning}.
\newblock In \emph{arXiv}, 2020.

\bibitem[Li et~al.(2019)Li, Yang, Zhou, and
  Hospedales]{Li2019Feature-criticGeneralization}
Yiying Li, Yongxin Yang, Wei Zhou, and Timothy~M Hospedales.
\newblock {Feature-critic networks for heterogeneous domain generalization}.
\newblock In \emph{ICML}, 2019.

\bibitem[Liang et~al.(2018)Liang, Li, and Srikant]{Liang2018EnhancingNetworks}
Shiyu Liang, Yixuan Li, and R~Srikant.
\newblock {Enhancing the reliability of out-of-distribution image detection in
  neural networks}.
\newblock In \emph{ICLR}, 2018.

\bibitem[Lin et~al.(2014)Lin, Chen, and Yan]{Lin2014NetworkNetwork}
Min Lin, Qiang Chen, and Shuicheng Yan.
\newblock {Network in network}.
\newblock In \emph{ICLR}, 2014.

\bibitem[Lin et~al.(2017)Lin, Goyal, Girshick, He, and
  Doll{\'{a}}r]{Lin2017FocalDetection}
Tsung-Yi Lin, Priya Goyal, Ross Girshick, Kaiming He, and Piotr Doll{\'{a}}r.
\newblock {Focal loss for dense object detection}.
\newblock In \emph{ICCV}, 2017.

\bibitem[Liu et~al.(2019)Liu, Borovykh, Grzelak, and
  Oosterlee]{Liu2019ACalibration}
Shuaiqiang Liu, Anastasia Borovykh, Lech~A Grzelak, and Cornelis~W Oosterlee.
\newblock {A neural network-based framework for financial model calibration}.
\newblock \emph{Journal of Mathematics in Industry}, 2019.

\bibitem[Lorraine et~al.(2020)Lorraine, Vicol, and
  Duvenaud]{Lorraine2020optimizingDifferentiation}
Jonathan Lorraine, Paul Vicol, and David Duvenaud.
\newblock {Optimizing millions of hyperparameters by implicit differentiation}.
\newblock In \emph{AISTATS}, 2020.

\bibitem[Luketina et~al.(2016)Luketina, Berglund, Klaus~Greff, and
  Raiko]{Luketina2016ScalableHyperparameters}
Jelena Luketina, Mathias Berglund, Aaltofi Klaus~Greff, and Tapani Raiko.
\newblock {Scalable gradient-based tuning of continuous regularization
  hyperparameters}.
\newblock In \emph{ICML}, 2016.

\bibitem[Minderer et~al.(2021)Minderer, Djolonga, Romijnders, Hubis, Zhai,
  Houlsby, Tran, and Lucic]{minderer2021revisiting}
Matthias Minderer, Josip Djolonga, Rob Romijnders, Frances Hubis, Xiaohua Zhai,
  Neil Houlsby, Dustin Tran, and Mario Lucic.
\newblock Revisiting the calibration of modern neural networks.
\newblock In \emph{NeurIPS}, 2021.

\bibitem[Mosser \& Naeini(2022)Mosser and Naeini]{mosser2022calibration}
Lukas Mosser and Ehsan~Zabihi Naeini.
\newblock A comprehensive study of calibration and uncertainty quantification
  for bayesian convolutional neural networks — an application to seismic
  data.
\newblock \emph{Geophysics}, 2022.

\bibitem[Mukhoti et~al.(2020)Mukhoti, Kulharia, Sanyal, Golodetz, Torr, and
  Dokania]{Mukhoti2020CalibratingLoss}
Jishnu Mukhoti, Viveka Kulharia, Amartya Sanyal, Stuart Golodetz, Philip H~S
  Torr, and Puneet~K Dokania.
\newblock {Calibrating deep neural networks using focal loss}.
\newblock In \emph{NeurIPS}, 2020.

\bibitem[M{\"{u}}ller et~al.(2019)M{\"{u}}ller, Kornblith, and
  Hinton]{Muller2019WhenHelp}
Rafael M{\"{u}}ller, Simon Kornblith, and Geoffrey Hinton.
\newblock {When does label smoothing help?}
\newblock In \emph{NeurIPS}, 2019.

\bibitem[Munir et~al.(2023)Munir, Khan, Khan, and Khan]{Munir2023BridgingPA}
Muhammad~Akhtar Munir, Muhammad~Haris Khan, Salman Khan, and Fahad~Shahbaz
  Khan.
\newblock Bridging precision and confidence: A train-time loss for calibrating
  object detection.
\newblock In \emph{CVPR}, 2023.

\bibitem[Naeini et~al.(2015)Naeini, Cooper, and
  Hauskrecht]{Naeini2015ObtainingBinning}
Pakdaman Naeini, Gregory~F Cooper, and Milos Hauskrecht.
\newblock {Obtaining well calibrated probabilities using bayesian binning}.
\newblock In \emph{AAAI}, 2015.

\bibitem[Netzer et~al.(2011)Netzer, Wang, Coates, Bissacco, Wu, and
  Ng]{Netzer2011ReadingLearning}
Yuval Netzer, Tao Wang, Adam Coates, Alessandro Bissacco, Bo~Wu, and Andrew~Y
  Ng.
\newblock {Reading digits in natural images with unsupervised feature
  learning}.
\newblock In \emph{NIPS Workshop on Deep Learning and Unsupervised Feature
  Learning}, 2011.

\bibitem[Ovadia et~al.(2019)Ovadia, Fertig, Ren, Nado, Sculley, Nowozin,
  Dillon, Lakshminarayanan, and Snoek]{ovadia2019trustUncertainty}
Yaniv Ovadia, Emily Fertig, Jie Ren, Zachary Nado, David Sculley, Sebastian
  Nowozin, Joshua~V Dillon, Balaji Lakshminarayanan, and Jasper Snoek.
\newblock Can you trust your model's uncertainty? evaluating predictive
  uncertainty under dataset shift.
\newblock In \emph{NeurIPS}, 2019.

\bibitem[Qin et~al.(2010)Qin, Liu, and Li]{Qin2010AMeasures}
Tao Qin, Tie-Yan Liu, and Hang Li.
\newblock {A general approximation framework for direct optimization of
  information retrieval measures}.
\newblock \emph{Information retrieval}, 2010.

\bibitem[Shu et~al.(2019)Shu, Xie, Yi, Zhao, Zhou, Xu, and
  Meng]{Shu2019Meta-Weight-Net:Weighting}
Jun Shu, Qi~Xie, Lixuan Yi, Qian Zhao, Sanping Zhou, Zongben Xu, and Deyu Meng.
\newblock {Meta-Weight-Net: learning an explicit mapping for sample weighting}.
\newblock In \emph{NeurIPS}, 2019.

\bibitem[Song et~al.(2023)Song, Wang, Mondal, and
  Sahoo]{Song2023AOpportunities}
Yisheng Song, Ting Wang, Subrota~K Mondal, and Jyoti~Prakash Sahoo.
\newblock {A comprehensive survey of few-shot learning: Evolution,
  applications, challenges, and opportunities}.
\newblock \emph{ACM Computing Surveys}, 2023.

\bibitem[Szegedy et~al.(2016)Szegedy, Vanhoucke, Ioffe, and
  Shlens]{Szegedy2016RethinkingVision}
Christian Szegedy, Vincent Vanhoucke, Sergey Ioffe, and Jon Shlens.
\newblock {Rethinking the Inception architecture for computer vision}.
\newblock In \emph{CVPR}, 2016.

\bibitem[Tomani et~al.(2021)Tomani, Gruber, Erdem, Cremers, and
  Buettner]{tomani2021calibrationDomain}
Christian Tomani, Sebastian Gruber, Muhammed~Ebrar Erdem, Daniel Cremers, and
  Florian Buettner.
\newblock Post-hoc uncertainty calibration for domain drift scenarios.
\newblock In \emph{CVPR}, 2021.

\bibitem[Vaicenavicius et~al.(2019)Vaicenavicius, Widmann, Andersson, Lindsten,
  Roll, and Sch\"{o}n]{vaicenavicius2019calibration}
Juozas Vaicenavicius, David Widmann, Carl Andersson, Fredrik Lindsten, Jacob
  Roll, and Thomas Sch\"{o}n.
\newblock Evaluating model calibration in classification.
\newblock In \emph{AISTATS}, 2019.

\bibitem[Wang et~al.(2021)Wang, Feng, and Zhang]{wang2021rethinking}
Deng-Bao Wang, Lei Feng, and Min-Ling Zhang.
\newblock Rethinking calibration of deep neural networks: do not be afraid of
  overconfidence.
\newblock In \emph{NeurIPS}, 2021.

\bibitem[Wang et~al.(2023)Wang, Ye, Liu, Dai, Kalander, Liu, Hao, and
  Han]{Wang2023OutofdistributionDW}
Qizhou Wang, Junjie Ye, Feng Liu, Quanyu Dai, Marcus Kalander, Tongliang Liu,
  Jianye Hao, and Bo~Han.
\newblock Out-of-distribution detection with implicit outlier transformation.
\newblock In \emph{ICLR}, 2023.

\bibitem[Wang et~al.(2018)Wang, Zhu, Torralba, and
  Efros]{Wang2018DatasetDistillation}
Tongzhou Wang, Jun-Yan Zhu, Antonio Torralba, and Alexei~A. Efros.
\newblock {Dataset distillation}.
\newblock In \emph{arXiv}, 2018.

\bibitem[Wang et~al.(2020)Wang, Yao, Kwok, and
  Ni]{Wang2020GeneralizingLearning}
Yaqing Wang, Quanming Yao, James Kwok, and Lionel~M. Ni.
\newblock {Generalizing from a few examples: A survey on few-shot learning}.
\newblock \emph{ACM Computing Surveys}, 2020.

\bibitem[Widmann et~al.(2019)Widmann, Lindsten, and
  Zachariah]{widmann2019calibration}
David Widmann, Fredrik Lindsten, and Dave Zachariah.
\newblock Calibration tests in multi-class classification: A unifying
  framework.
\newblock In \emph{NeurIPS}, 2019.

\bibitem[Wiseman(2022)]{wiseman2022cars}
Yair Wiseman.
\newblock Autonomous vehicles.
\newblock In \emph{Research Anthology on Cross-Disciplinary Designs and
  Applications of Automation}, 2022.

\bibitem[Yang et~al.(2018)Yang, Morillo, and Hospedales]{Yang2018DeepTrees}
Yongxin Yang, Irene~Garcia Morillo, and Timothy~M Hospedales.
\newblock {Deep neural decision trees}.
\newblock In \emph{ICML Workshop on Human Interpretability in Machine
  Learning}, 2018.

\bibitem[Yu et~al.(2023)Yu, Liu, and Wang]{Yu2023DatasetDA}
Ruonan Yu, Songhua Liu, and Xinchao Wang.
\newblock Dataset distillation: A comprehensive review.
\newblock \emph{ArXiv}, 2023.

\bibitem[Zagoruyko \& Komodakis(2016)Zagoruyko and
  Komodakis]{Zagoruyko2016WideNetworks}
Sergey Zagoruyko and Nikos Komodakis.
\newblock {Wide residual networks}.
\newblock In \emph{BMVC}, 2016.

\bibitem[Zhang et~al.(2021)Zhang, Marklund, Dhawan, Gupta, Levine, and
  Finn]{Zhang2021AdaptiveShift}
Marvin Zhang, Henrik Marklund, Nikita Dhawan, Abhishek Gupta, Sergey Levine,
  and Chelsea Finn.
\newblock {Adaptive risk minimization: learning to adapt to domain shift}.
\newblock In \emph{NeurIPS}, 2021.

\end{thebibliography}

\appendix

\section{Learned L2 regularization values}
\begin{figure}[h!]
\vskip 0.1in
\begin{center}
\centerline{\includegraphics[width=\columnwidth]{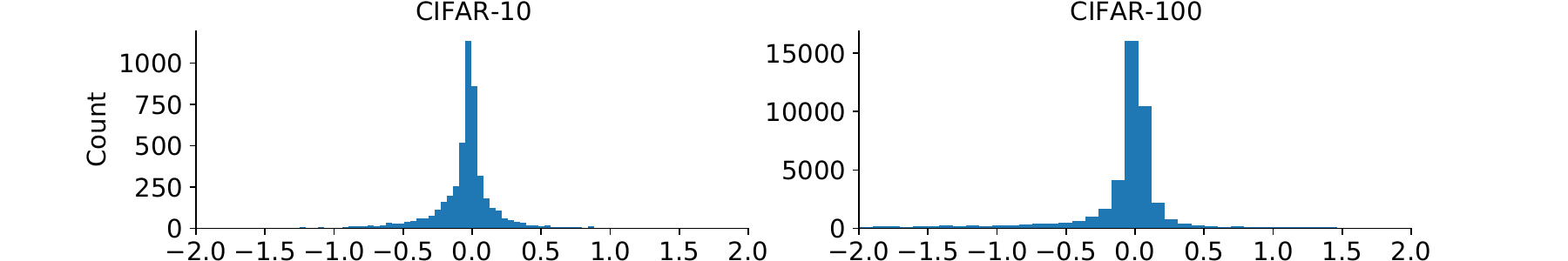}}
\caption{Histograms of the learned unit-wise L2 classifier regularization values.}
\label{fig:l2-analysis}
\end{center}
\vskip -0.1in
\end{figure}

We show the learned unit-wise L2 classifier regularization values in Figure \ref{fig:l2-analysis}. The results show there is a large variability in the coefficients and they take both positive and negative values.  

\section{Reliability analysis}
\begin{figure}[h!]
\vskip 0.1in
\begin{center}
\centerline{\includegraphics[width=\columnwidth]{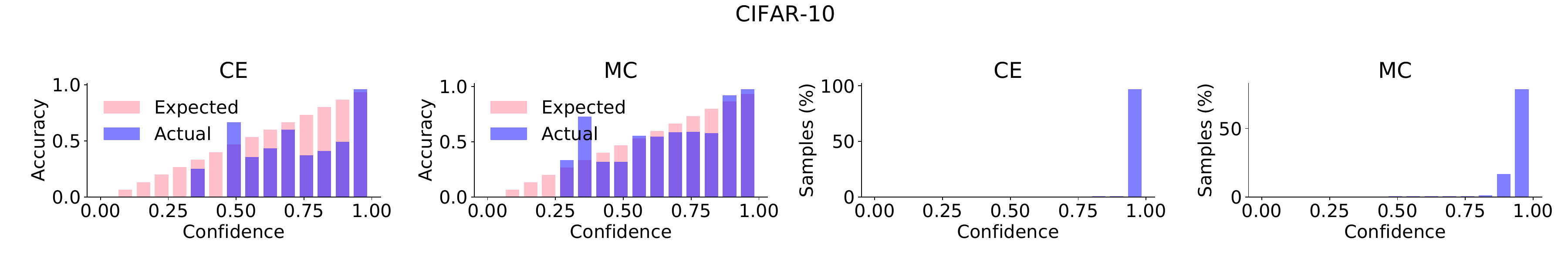}}
\centerline{\includegraphics[width=\columnwidth]{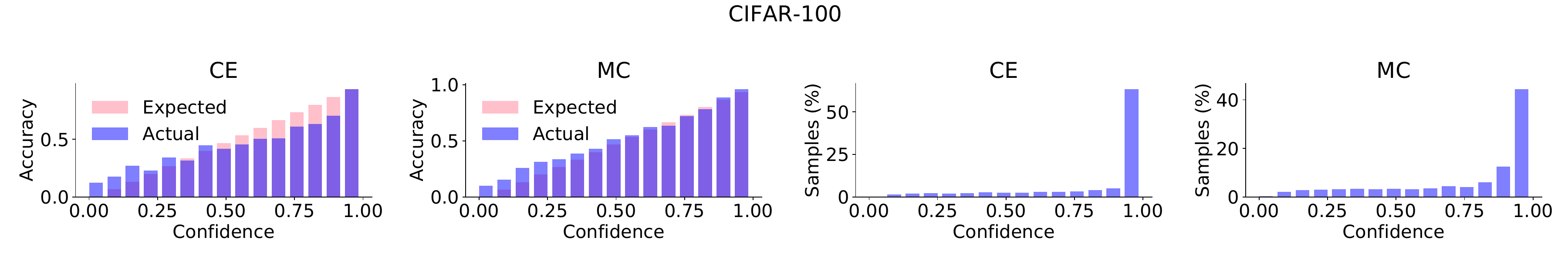}}
\caption{Reliability analysis for CIFAR-10 and CIFAR-100, using ResNet18 model.}
\label{fig:cifar100ece}
\end{center}
\vskip -0.1in
\end{figure}

We perform reliability analysis and show the percentage of samples with various confidence levels in Figure \ref{fig:cifar100ece}
for CIFAR-10 and CIFAR-100. We use ResNet18 and take the best model from training -- early stopping. The figure shows learnable label smoothing leads to visually better alignment between the expected and actual confidence binning. It also leads to softening the confidences of predictions, which is expected for label smoothing.

\section{Training times analysis}
We report the training times of the different approaches in Table \ref{tab:time} (for simplicity we only include CE baseline as all baselines have similar training time). We see that even if training with Meta-Calibration takes longer, the overall training time remains manageable. If excellent calibration is key, longer training time is acceptable.

\begin{table*}[h!]
\caption{Training times in hours. We used one NVIDIA Titan X for each experiment. Meta-Calibration takes longer, but if excellent calibration is a priority, the additional time is acceptable.}
\label{tab:time}
\vskip 0.1in
\centering
\resizebox{0.6\linewidth}{!}{\begin{tabular}{llcc}
\toprule
Dataset & Model & CE & MC (Ours) \\
\midrule
\multirow{4}{*}{CIFAR-10} & ResNet18 & \phantom{0}2.8h $\pm$ 0.1h & \phantom{0}7.0h $\pm$ 0.2h \\
& ResNet50 & \phantom{0}8.9h $\pm$ 0.1h & 21.9h $\pm$ 0.1h \\
& ResNet110 & 17.3h $\pm$ 0.3h & 44.6h $\pm$ 7.1h \\
& WideResNet26-10 & \phantom{0}9.2h $\pm$ 0.3h & 29.7h $\pm$ 0.2h \\
\midrule
\multirow{4}{*}{CIFAR-100} & ResNet18 & \phantom{0}2.7h $\pm$ 0.1h & 12.4h $\pm$ 0.2h \\
& ResNet50 & \phantom{0}8.9h $\pm$ 0.1h & 26.4h $\pm$ 0.2h \\
& ResNet110 & 17.3h $\pm$ 0.2h & 50.5h $\pm$ 1.0h \\
& WideResNet26-10 & \phantom{0}9.3h $\pm$ 0.4h & 39.4h $\pm$ 6.7h \\
\midrule
SVHN & ResNet18 & \phantom{0}4.4h $\pm$ 0.6h & \phantom{0}9.6h $\pm$ 0.1h \\
\midrule
20 Newsgroups & Global Pooling CNN & \phantom{0}0.1h $\pm$ 0.0h & \phantom{0}0.3h $\pm$ 0.1h \\
\bottomrule
\end{tabular}}
\vskip -0.1in
\end{table*}

\section{Cross-domain evaluation}

\paragraph{Setup}
To evaluate our framework's calibration performance under distribution shift, we use the CIFAR-C benchmark  \citep{Hendrycks2019BenchmarkingPerturbations}. CIFAR-C contains a relatively large variety of image corruptions, such as adding fog, pixelation, changes to the brightness and many others (overall 19 corruption types). There are 5 levels of severity for each of them, which can be interpreted as providing 95 domains. We select 22 of these as test domains following  \citep{Zhang2021AdaptiveShift}. These include impulse noise, motion blur, fog, and elastic transform at all levels of severity, and spatter and JPEG compression at the maximum level of severity. All other domains from \citep{Hendrycks2019BenchmarkingPerturbations} are used during training and validation.

All approaches are trained with augmented (corrupted) data so that they can better generalize across domains. In each step we sample a corruption to use from the set of training corruptions. Meta-Calibration in particular samples a separate corruption for the inner and outer loop so that it can train meta-parameters that are more likely to generalize to new domains.

\paragraph{Results}
We have evaluated our approach using CIFAR-C benchmark and ResNet18 model, and we show the results in Table \ref{tab:cross_domain_ece} and \ref{tab:cross_domain_err}. We include both the average across all test domains as well as the result on the most challenging domain (worst case). We report the mean and standard deviation across three repetitions. Meta-Calibration is still helpful for obtaining better calibration in the cross-domain case, but not to as large an extent as for clean data. While the efficacy of this algorithm likely depends on whether the domain-shifts seen during meta-training are representative in strength of those seen during meta-testing, we still view these as encouraging initial results that tools from multi-domain meta-learning can be adapted to address model calibration under domain shift. Studying how to more successfully exploit meta-learning for calibration in cross-domain scenarios is an interesting research question that the ML community can focus on in the future.

\begin{table*}[h!]
\caption{Test ECE (\%, $\downarrow$) for our cross-domain experiments on CIFAR-C.}
\label{tab:cross_domain_ece}
\centering
\resizebox{\linewidth}{!}{
\begin{tabular}{llcccccccccc}
\toprule
Dataset & Case & CE & LS & Brier & MMCE & FL & FLSD & MC (Ours) \\
\midrule
\multirow{2}{*}{CIFAR-10-C} & Average & \phantom{0}7.28 $\pm$ 0.03 & 3.13 $\pm$ 0.19 & 4.20 $\pm$ 0.81 & 4.12 $\pm$ 0.23 & 4.02 $\pm$ 0.26 & 4.13 $\pm$ 0.01 & \phantom{0}3.00 $\pm$ 0.28 \\
& Worst case & 15.55 $\pm$ 0.32 & 9.43 $\pm$ 0.65 & 9.19 $\pm$ 1.84 & 5.47 $\pm$ 0.78 & 4.97 $\pm$ 0.64 & 9.05 $\pm$ 0.88 & 10.97 $\pm$ 0.99 \\
\midrule
\multirow{2}{*}{CIFAR-100-C} & Average & \phantom{0}7.77 $\pm$ 0.09 & 5.13 $\pm$ 0.05 & 4.89 $\pm$ 0.06 & 6.07 $\pm$ 0.59 & 6.33 $\pm$ 0.23 & 7.06 $\pm$ 0.54 & \phantom{0}5.26 $\pm$ 2.03 \\
& Worst case & 14.75 $\pm$ 1.09 & 9.42 $\pm$ 0.71 & 7.40 $\pm$ 0.32 & 8.30 $\pm$ 0.75 & 8.53 $\pm$ 0.16 & 8.94 $\pm$ 0.61 & \phantom{0}6.79 $\pm$ 2.37 \\
\bottomrule
\end{tabular}}
\end{table*}

\begin{table*}[h!]
\caption{Test error (\%, $\downarrow$) for our cross-domain experiments on CIFAR-C.}
\label{tab:cross_domain_err}
\centering
\resizebox{\linewidth}{!}{
\begin{tabular}{llccccccccc}
\toprule
Dataset & Case & CE & LS & Brier & MMCE & FL & FLSD & MC (Ours) \\
\midrule
\multirow{2}{*}{CIFAR-10-C} & Average & 10.60 $\pm$ 0.05 & 10.55 $\pm$ 0.19 & 10.67 $\pm$ 0.11 & 11.08 $\pm$ 0.18 & 11.05 $\pm$ 0.15 & 10.13 $\pm$ 0.04 & 11.22 $\pm$ 0.22 \\
& Worst case & 21.83 $\pm$ 0.49 & 21.78 $\pm$ 0.18 & 21.32 $\pm$ 0.42 & 23.52 $\pm$ 0.29 & 23.34 $\pm$ 0.79 & 21.03 $\pm$ 0.80 & 23.57 $\pm$ 0.78 \\
\midrule
\multirow{2}{*}{CIFAR-100-C} & Average & 34.20 $\pm$ 0.09 & 35.26 $\pm$ 0.09 & 34.27 $\pm$ 0.23 & 34.26 $\pm$ 0.22 & 34.14 $\pm$ 0.38 & 33.32 $\pm$ 0.10 & 34.68 $\pm$ 0.13 \\
& Worst case & 57.10 $\pm$ 0.98 & 56.56 $\pm$ 0.91 & 56.95 $\pm$ 0.27 & 56.00 $\pm$ 0.20 & 55.87 $\pm$ 1.47 & 54.69 $\pm$ 0.44 & 55.76 $\pm$ 0.51 \\
\bottomrule
\end{tabular}}
\end{table*}

\section{Comparison to SB-ECE}
\begin{figure}[h!]
\vskip 0.1in
\begin{center}
\centerline{\includegraphics[width=\columnwidth]{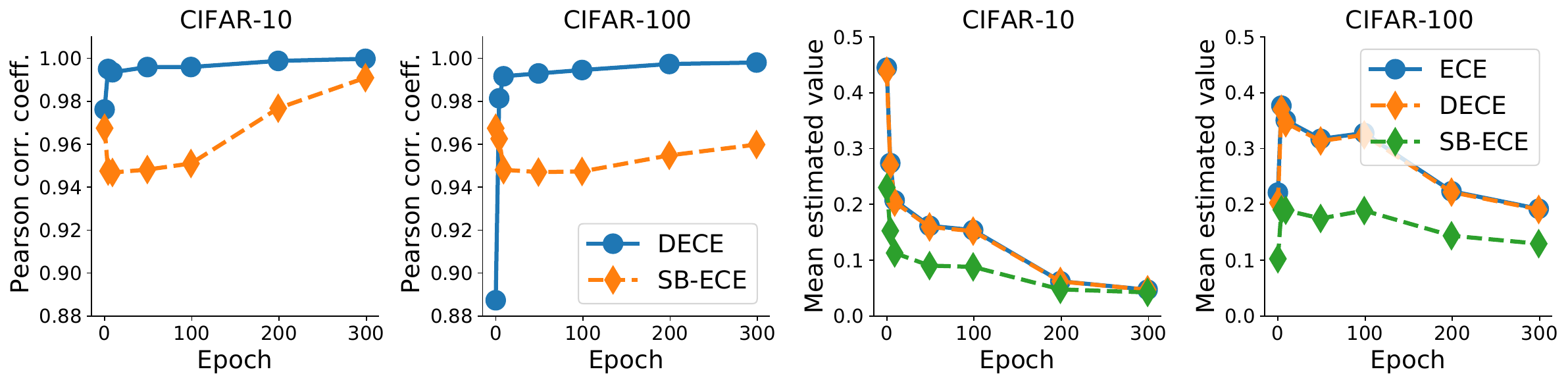}}
\caption{$T=0.01$: Pearson correlation coefficient between ECE/DECE and ECE/SB-ECE, and the mean estimated value of ECE, DECE and SB-ECE for CIFAR-10 and CIFAR-100 using ResNet18.}
\label{fig:sbece_analysis}
\end{center}
\vskip -0.1in
\end{figure}

\begin{figure}[h!]
\vskip 0.1in
\begin{center}
\centerline{\includegraphics[width=\columnwidth]{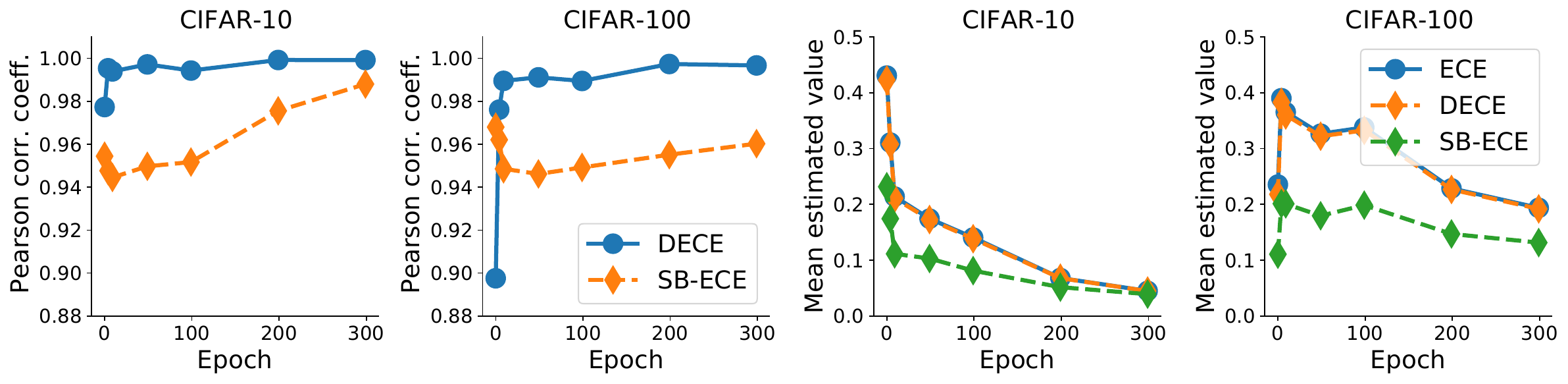}}
\caption{$T=0.0001$: Pearson correlation coefficient between ECE/DECE and ECE/SB-ECE, and the mean estimated value of ECE, DECE and SB-ECE for CIFAR-10 and CIFAR-100 using ResNet18.}
\label{fig:sbece_analysis2}
\end{center}
\vskip -0.1in
\end{figure}

Soft-binned ECE (SB-ECE) approximates the binning operation in ECE so that it is differentiable and can be used as an auxiliary loss during training \citep{Karandikar2021SoftNetworks}. Comparing DECE with SB-ECE, our DECE has both conceptual and empirical advantages. Conceptual advantages are as follows: 1) We also make the accuracy component of ECE differentiable, 2) SB-ECE binning estimate for the left-most and right-most bin can be inaccurate as a result of using bin's center value, while our binning approach does not suffer from this.
The empirical advantages are:
\begin{itemize}
    \item DECE provides a closer approximation to ECE than SB-ECE as empirically evaluated in Figure~\ref{fig:sbece_analysis} and \ref{fig:sbece_analysis2}. The quality of binning in SB-ECE is controlled by temperature parameter $T$, and we try both lower $T=0.0001$ and higher temperatures $T=0.01$. The results show DECE provides a significantly better approximation to ECE than SB-ECE regardless the value of the temperature. In fact, we see that the quality of SB-ECE approximation is relatively insensitive to the value of the temperature parameter $T$.
    \item Our Meta-Calibration with DECE leads to better calibration. \citet{Karandikar2021SoftNetworks} propose SB-ECE and SAvUC (soft version of accuracy versus uncertainty calibration loss \citep{krishnan2020improving}) to be used as auxiliary losses to encourage better calibration. We evaluate SB-ECE and SAvUC on our benchmark setup, both in their original form (as a regularizer added to CE or Focal Loss) and in our meta-learning framework (as part of meta-objective for LS learning) in Table~\ref{tab:sbece_results}. The results confirm the benefits of using Meta-Calibration with meta-objective that includes DECE. We additionally include comparison of error rates in Table \ref{tab:sbece_results_err}. The error rates remain similar to Meta-Calibration and the other baselines, although instabilities can occur that lead to noticeably worse error rates.
\end{itemize}

\citet{Karandikar2021SoftNetworks} also introduce an alternative way of training called interleaved training. In such training each epoch is split into two and a separate set is used for training with respect to calibration. We have implemented and evaluated interleaved training as described in \citep{Karandikar2021SoftNetworks}, and we report test ECE and error (\%) in Table \ref{tab:interleaved_ece} and \ref{tab:interleaved_err} respectively. The results suggest interleaved training generally leads to worse ECE than our approach, in most cases significantly worse. Interleaved training also leads to significantly worse error compared to our Meta-Calibration and the other baselines. We attribute the empirical benefits of Meta-Calibration to using the calibration objective for training only the label smoothing meta-parameters and also the specialised meta-training that uses an inner and outer loop.

\begin{table}[h!]
\caption{Test ECE (\%, $\downarrow$) -- comparison of Meta-Calibration (MC) that uses DECE vs SB-ECE and SAvUC.}
\label{tab:sbece_results}
\vskip 0.1in
\centering
\resizebox{1.\columnwidth}{!}{\begin{tabular}{llcccccccc}
\toprule
Dataset & Model & CE+SBECE & CE+SAvUC & FL3+SBECE & FL3+SAvUC & MC-SBECE & MC-SAvUC & MC (Ours)\\
\midrule
\multirow{2}{*}{CIFAR-10} & ResNet18 & \phantom{0}5.37 $\pm$ 0.16 & \phantom{0}3.26 $\pm$ 0.02 & \phantom{0}1.31 $\pm$ 0.01 & \phantom{0}1.91 $\pm$ 0.09 & \phantom{0}2.63 $\pm$ 0.66 & \phantom{0}4.18 $\pm$ 0.49 & 1.17 $\pm$ 0.26 \\
 & ResNet50 & \phantom{0}3.29 $\pm$ 0.37 & \phantom{0}3.36 $\pm$ 0.10 & \phantom{0}1.85 $\pm$ 0.03 & \phantom{0}1.54 $\pm$ 0.17 & \phantom{0}2.51 $\pm$ 0.90 & \phantom{0}2.50 $\pm$ 0.42 & 1.09 $\pm$ 0.09 \\
\midrule
\multirow{2}{*}{CIFAR-100} & ResNet18 & \phantom{0}5.21 $\pm$ 1.15 & \phantom{0}5.68 $\pm$ 0.26 & \phantom{0}2.77 $\pm$ 0.08 & \phantom{0}2.58 $\pm$ 0.10 & \phantom{0}5.72 $\pm$ 0.21 & \phantom{0}5.49 $\pm$ 0.54 & 2.52 $\pm$ 0.35 \\
 & ResNet50 & \phantom{0}5.72 $\pm$ 0.39 & 12.34 $\pm$ 0.09 & \phantom{0}5.86 $\pm$ 0.01 & \phantom{0}4.93 $\pm$ 0.18 & \phantom{0}3.10 $\pm$ 0.13 & \phantom{0}7.53 $\pm$ 0.63 & 3.07 $\pm$ 0.18 \\
\bottomrule
\end{tabular}}
\vskip -0.1in
\end{table}

\begin{table}[h!]
\caption{Test error (\%, $\downarrow$) -- comparison of Meta-Calibration (MC) that uses DECE vs SB-ECE and SAvUC.}
\label{tab:sbece_results_err}
\vskip 0.1in
\centering
\resizebox{\columnwidth}{!}{\begin{tabular}{llcccccccc}
\toprule
Dataset & Model & CE+SBECE & CE+SAvUC & FL3+SBECE & FL3+SAvUC & MC-SBECE & MC-SAvUC & MC (Ours)\\
\midrule
\multirow{2}{*}{CIFAR-10} & ResNet18 & \phantom{0}8.97 $\pm$ 0.04 & \phantom{0}5.07 $\pm$ 0.06 & \phantom{0}5.16 $\pm$ 0.10 & \phantom{0}5.20 $\pm$ 0.11 & \phantom{0}5.10 $\pm$ 0.12 & \phantom{0}5.17 $\pm$ 0.25 & 5.22 $\pm$ 0.06 \\
 & ResNet50 & 12.39 $\pm$ 0.39 & \phantom{0}5.03 $\pm$ 0.10 & \phantom{0}5.04 $\pm$ 0.10 & \phantom{0}5.26 $\pm$ 0.23 & \phantom{0}5.30 $\pm$ 0.05 & \phantom{0}5.16 $\pm$ 0.21 & 5.46 $\pm$ 0.05 \\
\midrule
\multirow{2}{*}{CIFAR-100} & ResNet18 & 27.55 $\pm$ 0.18 & 22.64 $\pm$ 0.32 & 22.87 $\pm$ 0.34 & 22.88 $\pm$ 0.26 & 23.98 $\pm$ 0.14 & 23.83 $\pm$ 0.25 & 23.88 $\pm$ 0.20 \\
 & ResNet50 & 28.33 $\pm$ 0.39 & 22.09 $\pm$ 0.13 & 22.40 $\pm$ 0.21 & 22.02 $\pm$ 0.41 & 23.51 $\pm$ 0.45 & 23.18 $\pm$ 0.26 & 23.22 $\pm$ 0.48 \\
\bottomrule
\end{tabular}}
\vskip -0.1in
\end{table}

\begin{table*}[h!]
\caption{Test ECE (\%, $\downarrow$) -- comparison of interleaved training and Meta-Calibration (MC).}
\label{tab:interleaved_ece}
\vskip 0.1in
\centering
\resizebox{0.7\linewidth}{!}{\begin{tabular}{llcccccc}
\toprule
Dataset & Model & CE+SBECE & CE+SAvUC & FL3+SBECE & FL3+SAvUC & MC (Ours)\\
\midrule
\multirow{2}{*}{CIFAR-10} & ResNet18 & \phantom{0}4.81 $\pm$ 1.25 & \phantom{0}4.67 $\pm$ 0.12 & \phantom{0}6.90 $\pm$ 0.18 & \phantom{0}1.52 $\pm$ 0.08 & 1.17 $\pm$ 0.26 \\
 & ResNet50 & \phantom{0}3.44 $\pm$ 0.37 & \phantom{0}4.55 $\pm$ 0.15 & \phantom{0}6.38 $\pm$ 0.82 & \phantom{0}1.38 $\pm$ 0.31 & 1.09 $\pm$ 0.09 \\
\midrule
\multirow{2}{*}{CIFAR-100} & ResNet18 & \phantom{0}5.17 $\pm$ 1.23 & 10.04 $\pm$ 0.65 & \phantom{0}4.81 $\pm$ 1.05 & \phantom{0}1.73 $\pm$ 0.24 & 2.52 $\pm$ 0.35 \\
 & ResNet50 & \phantom{0}6.74 $\pm$ 0.29 & 11.56 $\pm$ 2.98 & \phantom{0}4.35 $\pm$ 0.48 & \phantom{0}2.86 $\pm$ 0.27 & 3.07 $\pm$ 0.18 \\
\bottomrule
\end{tabular}}
\vskip -0.1in
\end{table*}

\begin{table*}[h!]
\caption{Test error (\%, $\downarrow$) -- comparison of interleaved training and Meta-Calibration (MC).}
\label{tab:interleaved_err}
\vskip 0.1in
\centering
\resizebox{0.7\linewidth}{!}{\begin{tabular}{llcccccccc}
\toprule
Dataset & Model & CE+SBECE & CE+SAvUC & FL3+SBECE & FL3+SAvUC & MC (Ours)\\
\midrule
\multirow{2}{*}{CIFAR-10} & ResNet18 & 19.09 $\pm$ 5.56 & \phantom{0}5.66 $\pm$ 0.13 & 14.22 $\pm$ 0.63 & \phantom{0}5.96 $\pm$ 0.13 & 5.22 $\pm$ 0.06 \\
 & ResNet50 & 13.51 $\pm$ 1.01 & \phantom{0}5.56 $\pm$ 0.13 & 12.63 $\pm$ 0.56 & \phantom{0}6.15 $\pm$ 0.45 & 5.46 $\pm$ 0.05 \\
\midrule
\multirow{2}{*}{CIFAR-100} & ResNet18 & 37.66 $\pm$ 2.85 & 26.94 $\pm$ 1.62 & 39.97 $\pm$ 5.71 & 26.64 $\pm$ 0.47 & 23.88 $\pm$ 0.20 \\
 & ResNet50 & 35.18 $\pm$ 1.24 & 24.75 $\pm$ 0.39 & 35.61 $\pm$ 2.22 & 25.45 $\pm$ 0.97 & 23.22 $\pm$ 0.48 \\
\bottomrule
\end{tabular}}
\vskip -0.1in
\end{table*}

\section{Comparison with the setup of \citet{Guo2017OnNetworks}}
Our main results adopt the setup from \citet{Mukhoti2020CalibratingLoss} as they provide a complete codebase and full details of their experiments. In this section we compare this setup with the one in \citep{Guo2017OnNetworks} that reports better ECE values in their paper. The comparison will focus on ResNet110 model trained with cross-entropy as it is the only model present in both our results and the results from \citet{Guo2017OnNetworks}.

We have tried to estimate the details of the training used by \citet{Guo2017OnNetworks}, based on the relatively limited information in the paper and the parameters used in the demo code for a different model. After estimating the hyper-parameters of \citep{Guo2017OnNetworks}, the differences compared to us are the following: \citet{Guo2017OnNetworks} train for 500 epochs and decay the learning rate by a factor of 10 after 250 and 375 epochs. In our main experiments we train for 350 epochs and decay the learning rate after 150 and 250 epochs. Minibatch size is 64 in \citep{Guo2017OnNetworks}, while we use 80 for ResNet110 (in general we use 128, but it was downscaled to 80 to fit into memory of our GPUs). We use weight decay of 5e-4 and no Nesterov momentum, while \citet{Guo2017OnNetworks} use Nesterov momentum and potentially use weight decay of 0.001. The demo code provides the value of weight decay, but it is then not used, so we have tried both including and excluding it. While the demo code uses 300 epochs, the paper says that 500 epochs are used for ResNet110, so we have used 500 epochs for the additional experiments. Further, the way that temperature scaling (TS) is done in \citep{Mukhoti2020CalibratingLoss} and \citep{Guo2017OnNetworks} is different. \citet{Mukhoti2020CalibratingLoss} use grid search over a range of values and argue that this gives stronger baselines than selecting the temperature by minimising validation set NLL. While we follow \citet{Mukhoti2020CalibratingLoss} in our paper, we also evaluate the alternative option. Based on the demo code, we try 50 iterations for optimising the temperature, but we also try 10 iterations because that value is mentioned in the paper of \citet{Guo2017OnNetworks}.

We report the results of our additional experiments in Table~\ref{tab:guo_comparison}. The results from the respective papers are at the top of each part, from which we observe that the stochastic depth model is better than the standard model. However, our comparison will focus on the standard model. Comparing \citep{Guo2017OnNetworks} and \citep{Mukhoti2020CalibratingLoss} we see that the test error in \citep{Guo2017OnNetworks} is significantly worse, as a result of the differences in the training and evaluation procedure. This is likely to relate to the general observation in \citep{Guo2017OnNetworks} that ``the network learns better classification accuracy at the expense of well-modelled probabilities''. The two papers report results from one run only, but for our experiments we report mean and standard deviation across three runs.

Our experiments that try to match the setup of \citet{Guo2017OnNetworks} show that grid-search based temperature scaling from \citep{Mukhoti2020CalibratingLoss} is significantly better, confirming the observations of \citet{Mukhoti2020CalibratingLoss}. We also observe that weight decay is beneficial, but while it gets us closer to the reported calibration, there is a significant difference in the error rate. Overall we see that neither of the configurations obtains similar values to the ones reported in \citep{Guo2017OnNetworks} for both CIFAR-10 and CIFAR-100 datasets and across all metrics. One of the configurations that we have evaluated ``With weight decay, grid search TS'' slightly improves over the setup from \citet{Mukhoti2020CalibratingLoss}, but it comes at an increased runtime cost (500 epochs instead of 350). To summarize, our additional comparison with the setup of \citet{Guo2017OnNetworks} has provided a more detailed justification for following the setup of \citet{Mukhoti2020CalibratingLoss} in our paper.

\begin{table*}[h!]
    \caption{Comparison of different setups for ResNet110 model. The results suggest neither configuration obtains similar results as reported in \citep{Guo2017OnNetworks} for both CIFAR-10 and CIFAR-100 datasets.}
    \label{tab:guo_comparison}
    \vskip 0.1in
    \centering
    \resizebox{1.\linewidth}{!}{\begin{tabular}{llccc}
    \toprule
    Dataset & Setup & Test ECE (\%, $\downarrow$) & Test ECE with TS (\%, $\downarrow$) & Test error (\%, $\downarrow$) \\
    \midrule
    \multirow{11}{*}{CIFAR-10} 
        & \citep{Guo2017OnNetworks} standard model & 4.60 & 0.83 & 6.21 \\
    & \citep{Guo2017OnNetworks} stochastic depth model & 4.12 & 0.60 & 5.64 \\
    & \citep{Mukhoti2020CalibratingLoss} standard model & 4.41 & 1.09 & 4.89 \\
    \cmidrule(l){2-5}
    & No weight decay, grid-search TS  & 5.96 $\pm$ 0.04 & 1.50 $\pm$ 0.24 & 6.39 $\pm$ 0.07 \\
    & No weight decay, optim. TS (50 iters) & 5.53 $\pm$ 0.12 & 5.16 $\pm$ 0.11 & 5.93 $\pm$ 0.14 \\
    & No weight decay, optim. TS (10 iters) & 5.53 $\pm$ 0.12 & 5.28 $\pm$ 0.12 & 5.93 $\pm$ 0.14 \\
    & With weight decay, grid-search TS & 4.04 $\pm$ 0.20 & 0.75 $\pm$ 0.13  & 4.56 $\pm$ 0.29 \\
    & With weight decay, optim. TS (50 iters) & 4.13 $\pm$ 0.12 & 3.59 $\pm$ 0.18 & 4.63 $\pm$ 0.06 \\
    & With weight decay, optim. TS (10 iters) & 4.13 $\pm$ 0.12 & 3.80 $\pm$ 0.16 & 4.63 $\pm$ 0.06 \\
    \cmidrule(l){2-5}
    & Our cross-entropy & 4.81 $\pm$ 0.12 &  1.49 $\pm$ 0.19 &  5.40 $\pm$ 0.10 \\
    & Our Meta-Calibration & 1.07 $\pm$ 0.12 &  1.33 $\pm$ 0.37 & 6.09 $\pm$ 0.22 \\
    \midrule
    \multirow{11}{*}{CIFAR-100} & \citep{Guo2017OnNetworks} standard model & 16.53 & 1.26 & 27.83 \\
    & \citep{Guo2017OnNetworks} stochastic depth model & 12.67 & 0.96 & 24.91 \\
    & \citep{Mukhoti2020CalibratingLoss} standard model & 19.05 & 4.43 & 22.73 \\
    \cmidrule(l){2-5}
    & No weight decay, grid-search TS  & 25.68 $\pm$ 0.50 & \phantom{0}1.85 $\pm$ 0.33 & 28.71 $\pm$ 0.55 \\
    & No weight decay, optim. TS (50 iters) & 25.88 $\pm$ 0.27 & 22.84 $\pm$ 0.22 & 29.24 $\pm$ 0.27 \\
    & No weight decay, optim. TS (10 iters) & 25.88 $\pm$ 0.27 & 23.96 $\pm$ 0.23 & 29.24 $\pm$ 0.27 \\
    & With weight decay, grid-search TS & 16.28 $\pm$ 0.22 & \phantom{0}3.54 $\pm$ 0.28 & 21.82 $\pm$ 0.19 \\
    & With weight decay, optim. TS (50 iters) & 16.73 $\pm$ 0.07 & 11.02 $\pm$ 0.20 & 22.17 $\pm$ 0.24 \\
    & With weight decay, optim. TS (10 iters) & 16.73 $\pm$ 0.07 & 12.86 $\pm$ 0.14 & 22.17 $\pm$ 0.24 \\
    \cmidrule(l){2-5}
    & Our cross-entropy & 14.96 $\pm$ 0.83 & \phantom{0}3.77 $\pm$ 0.51 &  22.99 $\pm$ 0.19 \\
    & Our Meta-Calibration & \phantom{0}2.80 $\pm$ 0.58 & \phantom{0}2.55 $\pm$ 0.33 & 24.51 $\pm$ 0.41 \\
    \bottomrule
    \end{tabular}}
    \vskip -0.1in
    \end{table*}

\end{document}